\documentclass[11pt,twoside]{article} 

\usepackage{amsmath,amsfonts,bm}

\def\eqref#1{equation~\ref{#1}}

\def\1{\bm{1}}

\DeclareMathAlphabet{\mathsfit}{\encodingdefault}{\sfdefault}{m}{sl}
\SetMathAlphabet{\mathsfit}{bold}{\encodingdefault}{\sfdefault}{bx}{n}

\DeclareMathOperator*{\argmin}{arg\,min}

\usepackage[utf8]{inputenc} 
\usepackage[T1]{fontenc}    
\usepackage{booktabs}       
\usepackage{amsfonts}       
\usepackage{nicefrac}       
\usepackage{microtype}      

\usepackage{subfigure}
\usepackage{epsf}
\usepackage{epsfig}
\usepackage{fancyhdr}
\usepackage{graphics}
\usepackage{graphicx}
\usepackage{psfrag}
\usepackage{fullpage}
\usepackage{pdfpages}
\usepackage{natbib}
\usepackage{url}
\usepackage[colorlinks,linkcolor=magenta,citecolor=blue, pagebackref=true]{hyperref}
\renewcommand*{\backrefalt}[4]{%
    \ifcase #1  {(Not cited.)}%
    \or         {(Cited on page~#2.)}%
    \else       {(Cited on pages~#2.)}%
    \fi}
\usepackage{color}

\usepackage{amsthm}
\usepackage{amsmath}
\usepackage{amssymb,bbm}
\usepackage{caption}
\usepackage{algorithm}
\usepackage{textcomp}
\usepackage{siunitx}
\usepackage{wrapfig}
\usepackage{multirow}
\usepackage{multicol}

\usepackage{caption}
\usepackage{algorithm}
\usepackage{algpseudocode}
\usepackage{amssymb}
\usepackage{graphicx}                        
\usepackage{tikz}
\definecolor{softblue}{RGB}{180,185,230}
\definecolor{softgreen}{RGB}{190,235,190}
\definecolor{softred}{RGB}{240,180,170}
\definecolor{boxgray}{RGB}{235,235,235}
\usetikzlibrary{positioning, arrows.meta}    
\usepackage{booktabs} 
\usepackage{multirow}  
\usepackage{graphicx}  
\usetikzlibrary{fit}

\newtheorem{assumption}{Assumption}

\newtheorem{theorem}{Theorem}
\newtheorem{proposition}{Proposition}
\newtheorem{definition}{Definition}

\def\argmin{\textnormal{arg} \min}

\begin{document}

\begin{center}

{\bf{\LARGE{Sliced-Regularized Optimal Transport
}}}
  
\vspace*{.2in}
{\large{
\begin{tabular}{c}
Khai Nguyen
\end{tabular}
}}

\vspace*{.2in}

\begin{tabular}{c}
Department of Statistics and Data Science \\
Texas A\&M University
\end{tabular}

\vspace*{.2in}
\today

\vspace*{.2in}

\begin{abstract}
We propose a new regularized optimal transport (OT) formulation, termed sliced-regularized optimal transport (SROT). Unlike entropic OT (EOT), which regularizes the transport plan toward an independent coupling, SROT regularizes it toward a sliced OT (SOT) plan. To the best of our knowledge, SROT is the first approach to leverage a version of SOT plan as a reference to improve classical OT. We provide a formal definition of SROT,  derive its dual formulation, and provide a post-Bayesian interpretation of SROT. { We then develop a Sinkhorn-style algorithm for efficient computation. It retains the scalability of EOT and, when the SOT plan is sparse, achieves even lower per-iteration complexity.}  By incorporating a scalable SOT plan as a prior, SROT yields more accurate approximations of the exact OT plan than EOT under the same level of regularization. Moreover, the resulting transport plan improves upon the reference SOT plan itself. We further introduce the corresponding OT divergence induced by SROT, named SROT divergence, and analyze its topological and computational properties. Finally, we validate our approach through experiments on synthetic datasets and color transfer tasks, demonstrating that SROT is better than both EOT and SOT in approximating exact OT. Additional experiments on gradient flows further highlight the advantages of SROT divergence\footnote{Code for this paper is published at  \url{https://github.com/khainb/SROT}.}.
\end{abstract}

\end{center}

 \section{Introduction}
\label{sec:introduction}
Optimal transport (OT)~\citep{villani2003topics,villani2009optimal} and Wasserstein distance are fundamental mathematical tools with various applications in statistics, machine learning, and data sciences. In generative modeling, it has been used to enhance  generative adversarial networks~\citep{arjovsky2017wasserstein,genevay2018learning},  flow-based models~\citep{lipman2023flow,pooladian2023multisample,tong2024improving}, and drifting models~\citep{he2026sinkhorn}. In addition, it is used to align source and target distributions in domain adaptation~\citep{courty2017joint,damodaran2018deepjdot}. Beyond these settings, it has found important applications in computational biology~\citep{bunne2023learning,schiebinger2019optimal} as well as in image processing~\citep{feydy2017optimal}, signal processing~\citep{kolouri2017optimal}, computer graphics~\citep{solomon2016entropic,solomon2015convolutional}, statistical inference~\citep{bernton2019approximate,bernton2019parameter,nguyen2026vertical}, dependency measurement~\citep{catalano2021measuring,catalano2024wasserstein}, among many others.

One challenge in OT is its computation. Entropic regularization~\citep{cuturi2013sinkhorn} is the most widely used and principled approach to accelerate OT via approximation. It smooths the transport problem by adding an entropy penalty, yielding a strongly convex formulation known as entropic OT (EOT). In the discrete setting, this regularization guarantees a unique optimal transport plan and enables efficient computation through matrix scaling algorithms such as Sinkhorn–Knopp~\citep{sinkhorn1967concerning}, or equivalently via iterative Bregman projections~\citep{benamou2015iterative}. Moreover, EOT gives rise to the Sinkhorn divergence~\citep{genevay2018learning}, which metrizes weak convergence of probability measures while remaining significantly more computationally efficient than the Wasserstein distance.  Beyond reducing computational complexity to near-quadratic time, entropic regularization improves statistical rates for estimating both the transport cost and Sinkhorn divergence~\citep{genevay2019sample} and the transport plan~\citep{manole2024plugin,rigollet2025sample}.

There are several extensions of entropic OT (EOT) obtained by modifying the regularization term. For example, quadratically regularized OT~\citep{lorenz2021quadratically} encourages diffuse but not overly entropic couplings by penalizing the squared mass of the transport plan. Sparsity-constrained OT~\citep{liu2023sparsityconstrained} enforces sparsity in the transportation plan, promoting more localized matchings. Low-rank OT~\citep{scetbon2022low} exploits a low-rank structure in the optimal plan to improve computational efficiency and scalability. While these regularizers encode useful inductive biases about desirable transportation patterns, they are typically non-informative with respect to the true optimal transport plan in the sense that they are not explicitly designed to favor couplings that are close to the unregularized OT solution. As a result, they may introduce bias away from the true OT geometry, potentially trading fidelity for computational or structural convenience. Moreover, recovering the true OT solution might require using very small regularization strengths, but this in turn can significantly degrade the computational advantages of the regularized formulations~\citep{altschuler2017near}, as it leads to slower convergence of numerical solvers.

Regularized OT methods can be viewed as encouraging the transport plan to remain close to a prescribed reference (prior) plan under a chosen notion of discrepancy, both of which may be specified explicitly or implicitly. On the discrepancy side, entropic OT (EOT)~\citep{cuturi2013sinkhorn} relies on the Kullback–Leibler (KL) divergence, more general Csiszár divergences are considered in~\citep{chizat2018scaling}, and quadratic regularization based on the $\mathbb{L}_2$ norm is studied in~\citep{lorenz2021quadratically}. Regarding the choice of prior, EOT adopts the independent coupling, while alternative works explore sparse plans~\citep{lorenz2021quadratically,liu2023sparsityconstrained}, low-rank structures~\citep{scetbon2022low}, and Gaussian couplings~\citep{freulon2025entropic}. As noted, these priors are typically non-informative for recovering the true optimal transport plan. However, designing an informative prior remains challenging, as the space of transport plans is highly complex and solving for the exact optimal plan is computationally demanding~\citep{peyre2020computational}.

We address this challenge by proposing to use a sliced OT (SOT)~\citep{rabin2012wasserstein,nguyen2025introduction} plan as the reference plan. SOT plans~\citep{mahey2023fast,liu2025expected,tanguy2025sliced,chapel2026differentiable} have recently emerged as efficient proxies for OT plans due to their favorable computational properties. The key idea of SOT is to project high-dimensional probability measures onto one-dimension, where the OT problem admits a closed-form solution, yielding a one-dimensional transport plan. These one-dimensional plans are then lifted back to the original space, and the final transport plan is obtained by aggregating over projections. While being computationally efficient, SOT plans provide only proxy to the true OT  and depend critically on the choice and design of projection functions~\citep{chapel2026differentiable,liu2025expected}. Nevertheless, using an SOT plan as a reference can yield a more informative prior, inducing a different structure in the resulting new regularized OT problem and potentially improving both the quality of the prior and the final plan.

For related work, one-dimensional Kantorovich potentials derived from SOT have been used to initialize the potentials in EOT~\citep{thornton2023rethinking}, although the optimization objective remains that of EOT. In amortized settings involving multiple pairs of probability measures, SOT transportation costs and Kantorovich potentials have also been leveraged to predict their (entropic) OT counterparts~\citep{nguyen2026fast,truong2026amortized}. However, to the best of our knowledge, no prior work has employed a SOT plan as a reference measure to define a new class of regularized OT problems. In summary, our contributions are threefold:

1. We propose \emph{sliced-regularized optimal transport} (SROT), a novel regularized OT formulation that employs a smoothed SOT plan as the reference measure and the KL divergence as the discrepancy. We establish the dual formulation of SROT and provide a post-Bayesian interpretation. Furthermore, we derive a Sinkhorn-type algorithm for SROT, closely mirroring the structure of EOT. {Finally, SROT admits a sparse implementation that can improve computation of the Sinkhorn algorithm.}

2. We introduce the SROT divergence, analogous to the Sinkhorn divergence in EOT. We show that this divergence is symmetric, non-negative, and discriminative. {In addition, we prove that it metrizes weak convergence of probability measures for a fixed reference plan.}

3. We validate our approach through experiments on synthetic datasets with diverse geometries, including Half Moons, Eight Gaussians, and Two Rings, demonstrating the effectiveness of SROT in recovering OT plans. We also perform extensive ablation studies on key hyperparameters, such as regularization strength, number of Sinkhorn iterations, and number of projections, highlighting the robustness of the method. These findings are further supported by a color transfer task. Finally, we present a gradient flow experiment illustrating the favorable behavior of the SROT divergence.

\textbf{Organization.} We begin by reviewing background on OT, EOT, and SOT in Section~\ref{sec:background}. Next, we discuss the definition of  SROT, its duality and computation, and SROT divergence in~\ref{sec:SROT}. Section~\ref{sec:experiments} presents discussed experiments. Finally, we conclude in Section~\ref{sec:conclusion}. Additional materials, including technical proofs and experimental results are provided in the Appendices.


\section{Background}
\label{sec:background}
We begin by reviewing the background on OT, EOT, and SOT, and introduce the necessary notation.

\textbf{Optimal Transport.} Given two probability measures $\mu \in \mathcal{P}(\mathcal{X})$ and $\nu\in \mathcal{P}(\mathcal{Y})$, and a ground metric $c:\mathcal{X}\times \mathcal{Y} \to \mathbb{R}_+$, the OT~\citep{villani2009optimal} problem is defined as follows:
$
    \pi^\star \in \argmin_{\pi \in \Pi(\mu,\nu)} \int_{\mathcal{X}\times \mathcal{Y}} c(x,y) \mathrm{d} \pi(x,y),
$
where $\Pi(\mu,\nu)$ is the set of admissible transportation plans (joint measures) between $\mu$ and $\nu$, and $\pi^\star$ is the optimal transportation plan. 

\textbf{Entropic-Regularized Optimal Transport.} EOT~\citep{cuturi2013sinkhorn} smoothens the OT problem and it can be conveniently written in a single convex optimization problem as follows:
\begin{align}
\label{eq:EOT}
    \pi^\star_\varepsilon = \argmin_{\pi \in \Pi(\mu,\nu)} \int_{\mathcal{X}\times \mathcal{Y}} c(x,y) \mathrm{d} \pi(x,y) +\varepsilon \text{KL}(\pi\mid \mu \otimes \nu),
\end{align}
where $\text{KL}(\pi\mid \xi) = \int_{\mathcal{X}\times \mathcal{Y}} \left(\log \left(\frac{\mathrm{d} \pi}{\mathrm{d} \xi}(x,y)\right)\right)\mathrm{d} \pi(x,y)$ and $\mu \otimes \nu$ is the product measure of $\mu$ and $\nu$. The optimal plan of EOT is unique as the problem is strongly convex. Let
 \begin{align}
\mathrm{OT}_{\varepsilon}(\mu,\nu)
=
\min_{\pi \in \Pi(\mu,\nu)} \int_{\mathcal{X}\times \mathcal{Y}} c(x,y) \mathrm{d} \pi(x,y) +\varepsilon \text{KL}(\pi\mid \mu \otimes \nu),
\end{align}
be the EOT functional, Sinkhorn divergence~\citep{genevay2018learning} is defined as follows:
\begin{align}
\mathcal{S}_{\varepsilon}(\mu,\nu)
=
\mathrm{OT}_{\varepsilon}(\mu,\nu)
-\frac{1}{2}\mathrm{OT}_{\varepsilon}(\mu,\mu)
-\frac{1}{2}\mathrm{OT}_{\varepsilon}(\nu,\nu).
\end{align}
Sinkhorn divergence helps to remove the bias created by entropic regularization. In particular, $\mathrm{OT}_{\varepsilon}(\mu,\nu)=0 \nRightarrow \mu =\nu$ but  $\mathcal{S}_{\varepsilon}(\mu,\nu)=0 \Rightarrow \mu=\nu$.


\textbf{Sliced Optimal Transport.} SOT considers a function $\mathbb{P}_\theta^c :\mathcal{X}\cup \mathcal{Y} \to \mathbb{R}$ where $c:\mathcal{X}\times \mathcal{Y} \to \mathbb{R}$ is the ground metric and  $\theta \sim \sigma(\theta) \in \mathcal{P}(\Theta)$ where $\Theta$ is the space of projection parameters. For example, when $c(x,y) =\|x-y\|_2$, we  have $\theta \sim \mathcal{U}(\mathbb{S}^{d-1})$ and $\mathbb{P}_\theta^c  =\langle \theta,x\rangle$~\citep{bonneel2015sliced,rabin2012wasserstein}. For other geometry, we might need to use other types of projections~\citep{kolouri2019generalized,bonet2024sliced,nguyen2026summarizing}.  We note that there might not always be a mapping from $c$ to $\mathbb{P}_\theta^c$ as designing projection function for SOT is still an active area of research. For $\mu \in \mathcal{P}(\mathcal{X})$ and $\nu \in \mathcal{P}(\mathcal{Y})$, the one-dimensional OT plan with $\mathbb{P}_\theta^c$ admits the following closed-form:
$
\underline{\pi}_\theta = (F_{\mathbb{P}_\theta^c\sharp \mu}^{-1},F_{\mathbb{P}_\theta^c\sharp \nu}^{-1})\sharp \mathcal{U}([0,1]),
$
where $F_{\mathbb{P}_\theta^c\sharp \mu}^{-1}$ and $F_{\mathbb{P}_\theta^c\sharp \nu}^{-1}$ are quantile functions respectively. With  the one-dimensional OT plan $\underline{\pi}_\theta$, we can construct a lifted transportation plan~\citep{muzellec2019subspace,tanguy2025sliced} as follows: 
\begin{align}
\label{eq:liftedplan}
    \pi_\theta =  \mu_{t_1} \otimes \nu_{t_2} \otimes \underline{\pi}_\theta,
\end{align}
where $\mu$ and $\nu$ are disintegrated as $\mu_{t_1} \otimes \nu_{t_2}$ with respect to $\underline{\pi}_\theta(t_1,t_2)$. In particular, we can write out
  $    \int_{\mathcal{X}\times \mathcal{Y}} c(x,y) \mathrm{d}  \pi_\theta(x,y) = \int_{\mathbb{R} \times \mathbb{R}} \int_{ (\mathbb{P}_{\theta}^{c})^{-1}(t_1) \times (\mathbb{P}_{\theta}^{c})^{-1}(t_2)} c(x,y) \mathrm{d}\mu_{t_1} \otimes \nu_{t_2}(x,y) \mathrm{d} \underline{\pi}_\theta(t_1,t_2)   ,
$
which is the transportation cost of $\pi_\theta$.   The final transportation plan from SOT is then defined by averaging over all $\theta \sim \sigma(\theta)$~\citep{liu2025expected}:
\begin{align}
\label{eq:SOT}
    \pi^{\mathrm{SOT}} = \mathbb{E}_{\theta \sim \sigma(\theta)}[\pi_\theta],
\end{align}
where $\sigma$ can be chosen to be uniform~\citep{rowland2019orthogonal} (with numerical approximation), Softmin of transportation cost of ${\pi}_\theta$ for $\theta \in \{\theta_1,\ldots,\theta_L\}$ ($L\geq 2$)~\citep{liu2025expected}, and searching for ${\pi}_\theta$ with the minimum cost~\citep{mahey2023fast,chapel2026differentiable}.

\textbf{Discrete Cases.}  In practice, we often work with discrete probability measures. In particular, we have $\mu= \sum_{i=1}^n\alpha_i \delta_{x_i}$ and $\nu= \sum_{j=1}^m \beta_j \delta_{y_j}$ with $\sum_{i=1}^n \alpha_i = \sum_{j=1}^m \beta_j=1$ and $\alpha_i>0,\beta_j>0 \forall i,j$. In this case, the EOT problem becomes:
\begin{align}
    P^\star_\varepsilon = \argmin_{P \in \Gamma(\boldsymbol{\alpha},\boldsymbol{\beta})}\langle C,P\rangle +\varepsilon \text{KL}(P\mid P_0),
\end{align}
where $\Gamma(\boldsymbol{\alpha},\boldsymbol{\beta}) = \{P \in \mathbb{R}_+^{n \times m} \mid P\mathbf{1} = \boldsymbol{\alpha}, P^\top \mathbf{1} = \boldsymbol{\beta}\}$ is the set of discrete plans,  $P_0 = \boldsymbol{\alpha} \boldsymbol{\beta}^\top$,  $  \text{KL}(P\mid P_0)= \sum_{i=1}^n \sum_{j=1}^m P_{ij}\log \left(\frac{P_{ij}}{P_{0,ij}}\right)$, and $C_{ij}=c(x_i,y_j)$. In this case, the SOT plan, denoted as $P_\theta^{SOT}$~\eqref{eq:liftedplan} and $P^{\mathrm{SOT}}$~\eqref{eq:SOT} can be obtained efficiently in a closed-form~\citep{liu2025expected} based on sorting permutation of atoms of $\mathbb{P}_\theta^c \sharp \mu =\sum_{i=1}^n\alpha_i \delta_{\mathbb{P}_\theta^c(x_i)}$ and $\mathbb{P}_\theta^c \sharp \nu =\sum_{j=1}^m \beta_j \delta_{\mathbb{P}_\theta^c(y_j)}$ respectively. The key computational benefit comes from the fact that sorting only costs $\mathcal{O}(n \log n)$ and $\mathcal{O}(m \log m)$.

\section{Sliced-Regularized Optimal Transport}
\label{sec:SROT}
In this section, we define primal and dual formulation of SROT in Section~\ref{subsec:primal_dual_form}. We then propose the computational algorithm of SROT in Section~\ref{subsec:computation_algorithms}. Finally, we introduce SROT divergence  and discuss its topological properties in Section~\ref{subsec:SROTdivergence}.

\subsection{Primal and Dual Formulation}
\label{subsec:primal_dual_form}

\begin{figure}[!t]
    \centering
    \begin{tabular}{cc}
    \includegraphics[width=0.45\linewidth]{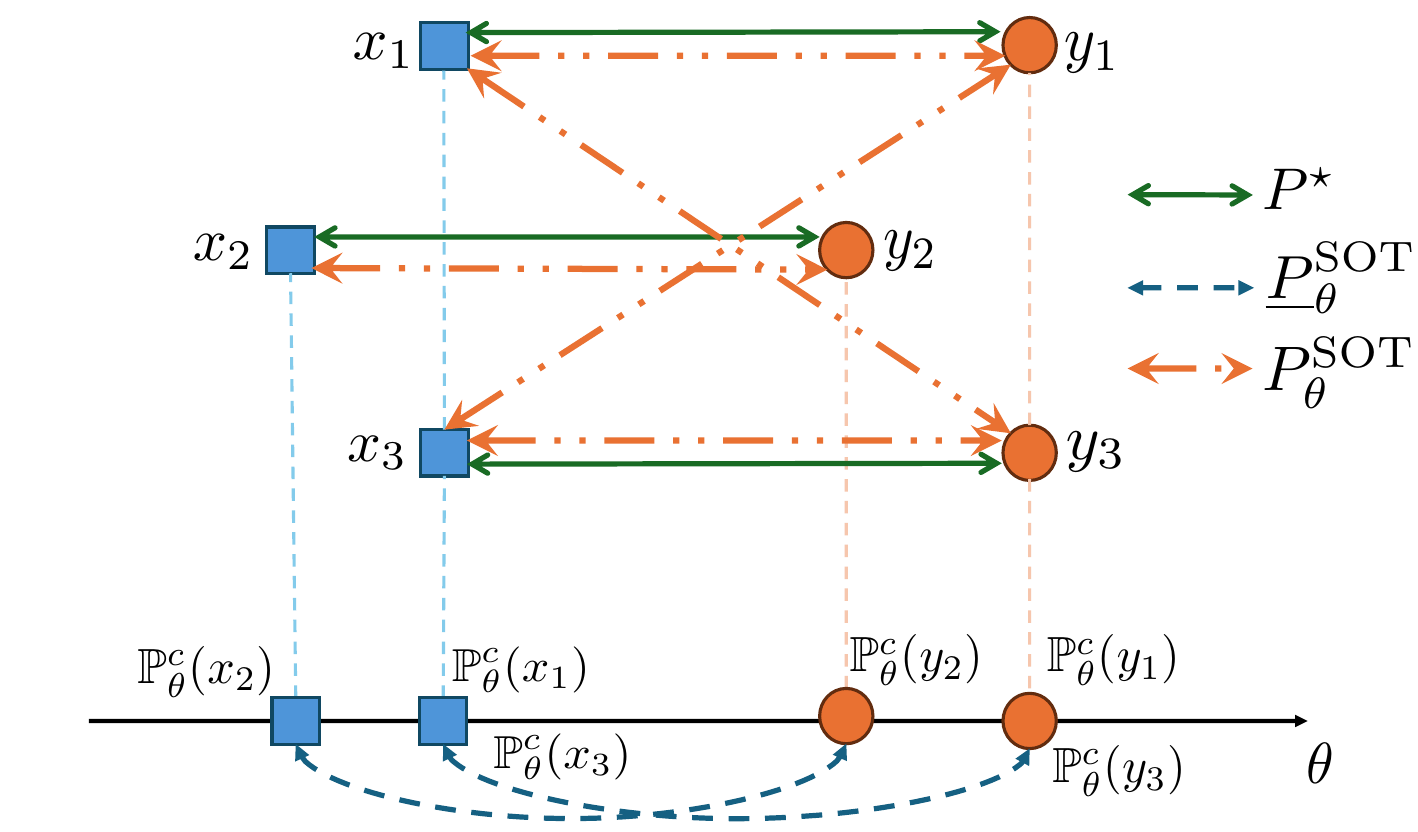} &  \includegraphics[width=0.4\linewidth]{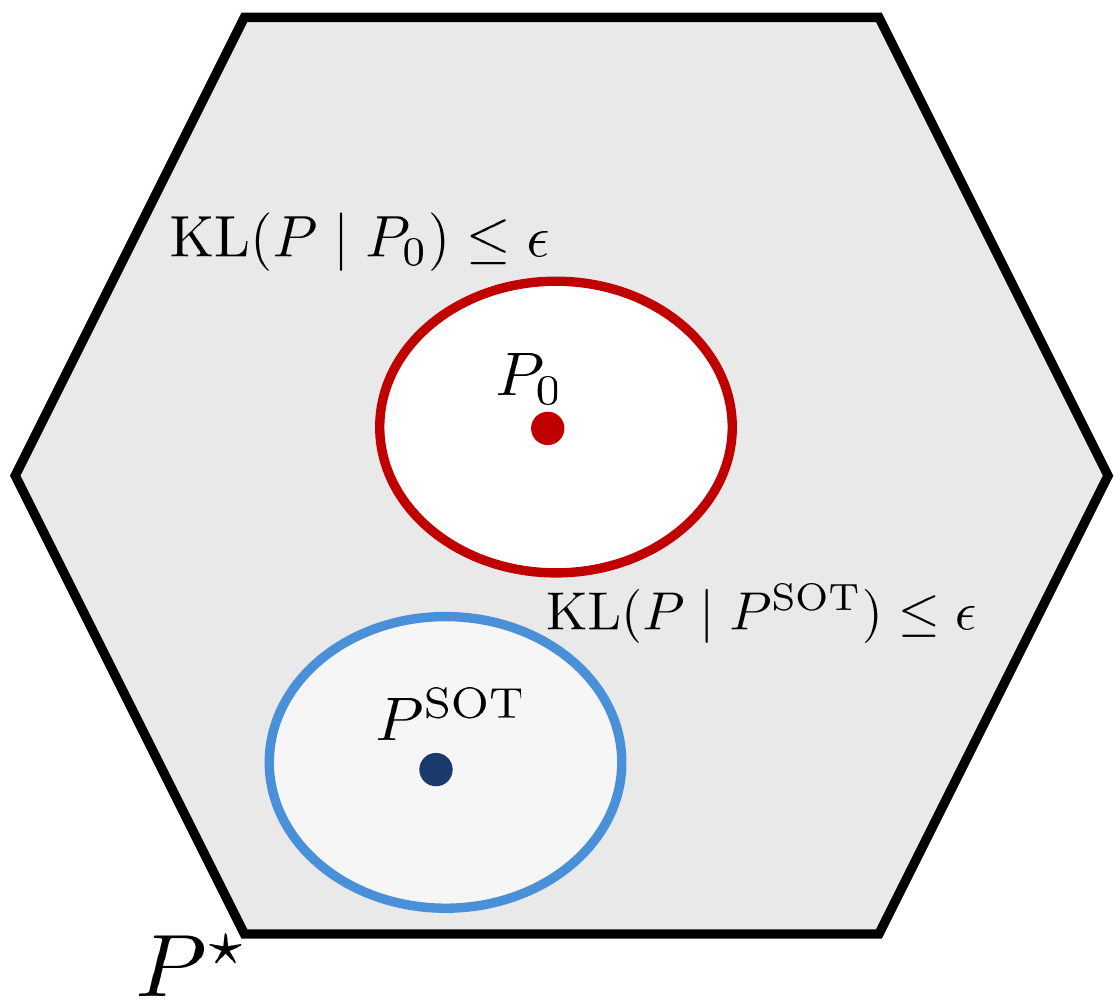} \\
       (a) An example of SOT plan & (b) Intuition of SROT 
    \end{tabular}
     \vspace{-0.5em}
    \caption{(a) An example of SOT plan (adapting from~\citep{nguyen2025introduction}) and (b) intuition of SROT: restricting the feasible set of plans  around $P^{\mathrm{SOT}}$ (there is a correspondence between $\varepsilon$ and  $\varepsilon$).} 
    \label{fig:SROT}
    \vspace{-1em}
\end{figure}

\begin{definition}[SROT]
Let $\mu \in \mathcal{P}(\mathcal{X})$ and $\nu \in \mathcal{P}(\mathcal{Y})$ be two probability
measures, $c : \mathcal{X} \times \mathcal{Y} \to \mathbb{R}_+$ a ground cost, 
$\varepsilon > 0$ a regularization parameter, the \emph{sliced-regularized optimal transport} (SROT) is defined as follows:
\begin{align}
  \label{eq:primal_cont}
  \pi^\star_{\varepsilon,\mathrm{SOT}} =
    \argmin_{\pi \in \Pi(\mu,\nu)}
      \int_{\mathcal{X}\times\mathcal{Y}} c(x,y)\mathrm{d}\pi(x,y)
      +
      \varepsilon\mathrm{KL}\left(\pi \mid \pi^{\mathrm{SOT}}\right),
\end{align}
where $\Pi(\mu,\nu)$ is the set of admissible transportation plans and  $ \pi^{\mathrm{SOT}}$ is the SOT reference plan.
\end{definition}
Compared to EOT in~\eqref{eq:EOT}, the reference coupling is replaced from $\mu \otimes \nu$ to $\pi^{\mathrm{SOT}}$. As $\varepsilon \to 0$, the optimal plan satisfies $\pi^\star_{\varepsilon,\mathrm{SOT}} \to \pi^\star$, whereas as $\varepsilon \to \infty$, we obtain $\pi^\star_{\varepsilon,\mathrm{SOT}} \to \pi^{\mathrm{SOT}}$. {Problem~\eqref{eq:primal_cont} is well posed for any SOT plan. In particular, the feasible set $\{\pi \in \Pi(\mu,\nu) : \pi \ll \pi^{\mathrm{SOT}}\}$ is non-empty since it contains $\pi^{\mathrm{SOT}}$ itself. Moreover, the objective is strictly convex, hence, the minimizer exists and is unique. Its support is contained in that of $\pi^{\mathrm{SOT}}$. To recover the exact OT plan as $\varepsilon \to 0$ and for the topological results of Section~\ref{subsec:SROTdivergence}, we need $\pi^{\mathrm{SOT}}$ to have full support on $\mathcal{X}\times\mathcal{Y}$, which is ensured by the smoothed reference plan $\pi^{\mathrm{SOT}}_\gamma = (1-\gamma)\pi^{\mathrm{SOT}} + \gamma\mu\otimes\nu$ for any $\gamma \in (0,1]$. In practice, we find that $\gamma = 0$ works well in approximating OT. To keep theory and practice consistent we use a very small $\gamma$, e.g., $\gamma = 10^{-8}$.} For convenience, we omit $\gamma$ from the notation and write $\pi^{\mathrm{SOT}}_\gamma$ simply as $\pi^{\mathrm{SOT}}$, with the understanding that a “SOT plan” may refer to its smoothed version when needed. When $\mu= \sum_{i=1}^n\alpha_i \delta_{x_i}$ and $\nu= \sum_{j=1}^m \beta_j \delta_{y_j}$ with $\sum_{i=1}^n \alpha_i = \sum_{j=1}^m \beta_j$,  SROT problem becomes:
\begin{align}
    P^\star_{\varepsilon,\mathrm{SOT}} = \argmin_{P \in \Gamma(\boldsymbol{\alpha},\boldsymbol{\beta})}\langle C,P\rangle +\varepsilon \sum_{i=1}^n\sum_{j=1}^m P_{ij}\log\left(\frac{P_{ij}}{P_{ij}^{\mathrm{SOT}}} \right),
\end{align}
where $\Gamma(\boldsymbol{\alpha},\boldsymbol{\beta})$ is the set of discrete plans and $C_{ij}=c(x_i,y_j)$.  We demonstrate  an example of a SOT plan in Figure~\ref{fig:SROT}(a) and the intuition of SROT in Figure~\ref{fig:SROT}(b) which is changing the center of feasible set of plans from independent plan to a SOT plan.

\textbf{A Post-Bayesian Interpretation.} To further explain why changing the reference plan is a natural idea, we discuss a post-Bayesian view of SROT. We consider a pair of random variables $(X,Y)$ representing a probabilistic alignment on $\mathcal{X} \times \mathcal{Y}$, with $X \sim \mu \in \mathcal{P}(\mathcal{X})$ and $Y \sim \nu \in \mathcal{P}(\mathcal{Y})$. We introduce a prior $\pi_0 \in \Pi(\mu,\nu)$ on $(X,Y)$ as our initial belief. After observing a ground metric $c$, we aim to update this belief. Following the generalized Bayes framework~\citep{bissiri2016general}, noting that specifying a sampling model for the function $c$ is non-trivial, we define a generalized likelihood as $\ell(c\mid X,Y) = \exp\left(-\frac{c(X,Y)}{\varepsilon}\right)$. The resulting generalized posterior is given by $\pi(X,Y\mid c) \propto \exp\left(-\frac{c(X,Y)}{\varepsilon}\right)\pi_0(X,Y)$. As shown in~\citep{bissiri2016general}, this posterior is the solution to the following optimization problem:
$
\argmin_{\pi \in \Pi(\mu,\nu)}
\int_{\mathcal{X}\times\mathcal{Y}} c(x,y),\mathrm{d}\pi(x,y)
+
\varepsilon\mathrm{KL}\left(\pi \mid \pi_0\right),
$
which is exactly the entropic OT problem with KL regularization and reference measure $\pi_0$. Therefore, when we use $\pi^{\mathrm{SOT}}$ as the reference plan, we effectively change the prior in this generalized Bayesian model. With a more informative prior, inference can become more concentrated. In SROT, the prior $\pi^{\mathrm{SOT}}$ depends implicitly on the observation $c$, making the approach an instance of empirical Bayes~\citep{robbins1992empirical}, i.e., a data-dependent prior. While we can use any transportation plans as the prior plan including low-rank OT~\citep{scetbon2022low}, SOT plans are fast to obtain. SROT serves as a bridge between the SOT literature and regularized OT. As the quality of SOT plans improves, the quality of the corresponding SROT plans also improves.

\textbf{Duality.} We now discuss the duality of SROT which we later use to derive our computational algorithms. Similar to EOT, SROT provides a strong duality.
\begin{theorem}[Duality  of SROT]
\label{theorem:dual_cont}
The strong dual of the minimization problem in \eqref{eq:primal_cont} is
\begin{align}
  \label{eq:dual_cont}
  \max_{f \in \mathcal{C}(\mathcal{X}), g \in \mathcal{C}(\mathcal{Y})}
  &\int_\mathcal{X} f(x)\mathrm{d}\mu(x)
  + \int_\mathcal{Y} g(y)\mathrm{d}\nu(y) \nonumber
   \\
   &- \varepsilon
    \int_{\mathcal{X}\times\mathcal{Y}}
      \exp\left(\frac{f(x)+g(y)-c(x,y)}{\varepsilon}\right)
    \mathrm{d}\pi^{\mathrm{SOT}}(x,y)
\end{align}
where $\mathcal{C}(\mathcal{X})$ and $\mathcal{C}(\mathcal{Y})$ are sets of continuous functions on $\mathcal{X}$ and $\mathcal{Y}$ respectively.
The optimal transport plan can be recovered as follows:
\begin{equation}
  \label{eq:primal_recovery}
  \mathrm{d}\pi^\star_\varepsilon(x,y)
  =
  \exp\left(\frac{f^\star(x) + g^\star(y) - c(x,y)}{\varepsilon}\right)
  \mathrm{d}\pi^{\mathrm{SOT}}(x,y),
\end{equation}
where  $(f^\star, g^\star)$  are optimal dual potentials.
\end{theorem}
The proof of Theorem~\ref{theorem:dual_cont} is given in Appendix~\ref{subsec:proof:theorem:dual_cont}.  Compared to EOT, the optimal plan of SROT is integrated  $\left(\frac{f^\star(x) + g^\star(y) - c(x,y)}{\varepsilon}\right)$ with respect to $\pi^{\mathrm{SOT}}(x,y)$ instead of $\mathrm{d}\mu(x) \mathrm{d} \nu(y)$. In the discrete case, 
$\mu = \sum_{i=1}^n \alpha_i \delta_{x_i}$ and
$\nu = \sum_{j=1}^m \beta_j \delta_{y_j}$, 
the dual problem reads
\begin{align}
\label{eq:discrete_dual}
\max_{\mathbf{f}\in \mathbb{R}^n,\mathbf{g} \in \mathbb{R}^m}
\langle \mathbf{f},\boldsymbol{\alpha}\rangle
+ \langle \mathbf{g},\boldsymbol{\beta}\rangle
- \varepsilon \sum_{i=1}^n \sum_{j=1}^m
P^{\mathrm{SOT}}_{ij}
\exp\left(\frac{\mathbf{f}_i+\mathbf{g}_j - C_{ij}}{\varepsilon}\right).
\end{align}
{We define $C_{ij}' = C_{ij} - \varepsilon \log P^{\mathrm{SOT}}_{ij}$. Since $0 < P^{\mathrm{SOT}}_{ij} \le \alpha_i$, the term $-\varepsilon\log P^{\mathrm{SOT}}_{ij}$ is positive for every pair and smallest for pairs with high probability under $P^{\mathrm{SOT}}$. Because adding $\varepsilon\log\alpha_i$ to the $i$-th row of the cost does not change the optimal plan, SROT is equivalent to EOT with the adjusted cost $C_{ij} + \varepsilon \log\big(\alpha_i / P^{\mathrm{SOT}}_{ij}\big)$, which penalizes pairs with low probability under $P^{\mathrm{SOT}}$ by $\varepsilon\log(\alpha_i/P^{\mathrm{SOT}}_{ij}) \ge 0$ (for uniform weights, $\varepsilon \log(1/(nP^{\mathrm{SOT}}_{ij}))$).}
An optimal coupling is recovered as
$
 P^\star_{\varepsilon,\mathrm{SOT},ij}
=
P^{\mathrm{SOT}}_{ij}
\exp\left(\frac{\mathbf{f}_i^\star+\mathbf{g}_j^\star - C_{ij}}{\varepsilon}\right),
$
where $(\mathbf{f}^\star,\mathbf{g}^\star)$ are optimal discrete potentials. We can see SROT with Sinkhorn solver as an automatic adjustment of the regularization strength $\varepsilon$ for each pair of atoms. We recall that SROT is a completely new formulation of regularized OT which can admit different solvers and can be generalized beyond KL divergence regularization.

{\begin{proposition}[Approximation guarantee]
\label{prop:approx}
Let $P_0 \in \Gamma(\boldsymbol{\alpha},\boldsymbol{\beta})$ have positive entries, let $P_\varepsilon$ be the minimizer of $\langle C,P\rangle + \varepsilon\mathrm{KL}(P\mid P_0)$ over $\Gamma(\boldsymbol{\alpha},\boldsymbol{\beta})$, and let $P^\star$ be an optimal plan of the unregularized problem. Then
$\langle C, P_\varepsilon\rangle - \langle C, P^\star\rangle \le \varepsilon\mathrm{KL}(P^\star \mid P_0)$.
\end{proposition}
Proposition~\ref{prop:approx} is the argument of~\cite{altschuler2017near} for a general reference plan. The proof is given in Appendix~\ref{subsec:proof:prop:approx}. For SROT the constant is $\mathrm{KL}(P^\star\mid P^{\mathrm{SOT}}_\gamma)$, which is small whenever the SOT plan puts mass on the pairs used by $P^\star$. This quantifies why a more informative reference yields a better approximation at the same $\varepsilon$.
}

\begin{figure}[!t]
\centering
\begin{minipage}{0.48\textwidth}
\begin{algorithm}[H]
\caption{SOT Plan}
\label{alg:SOT}
\begin{algorithmic}[1]
\Require Projection function $\mathbb{P}_\theta^c$
\State Initialize $P^{\mathrm{SOT}} \gets 0$
\For{$l = 1, \dots, L$}
    \State Sample $\theta_l \sim \mathcal{U}(\mathbb{S}^{d-1})$
    \State Obtain $P^{\mathrm{SOT}}_{\theta_l}$ and weights $w_l$
\EndFor
 \State $P^{\mathrm{SOT}}  \gets \sum_{l=1}^L \frac{w_l}{\sum_{l'=1}^L w_{l'}} P^{\mathrm{SOT}}_{\theta_l}$
\State \Return $P^{\mathrm{SOT}} $
\end{algorithmic}
\end{algorithm}
\end{minipage}
\begin{minipage}{0.48\textwidth}
\begin{algorithm}[H]
\caption{SROT Plan}
\label{alg:SROT}
\begin{algorithmic}[1]
\Require $C$, $P^{\mathrm{SOT}}$, $\boldsymbol{\alpha}$, $\boldsymbol{\beta}$, $\varepsilon$
\State $K \gets P^{\mathrm{SOT}} \odot \exp(-C/\varepsilon)$
\State Initialize $\mathbf{u} \gets \mathbf{1}, \mathbf{v} \gets \mathbf{1}$
\While{not converged}
    \State $\mathbf{u} \gets \boldsymbol{\alpha} \oslash (K \mathbf{v})$
    \State $\mathbf{v} \gets \boldsymbol{\beta} \oslash (K^\top \mathbf{u})$
\EndWhile
\State \Return $P_{\varepsilon,\mathrm{SOT}} = \operatorname{diag}(\mathbf{u}) K \operatorname{diag}(\mathbf{v})$
\end{algorithmic}
\end{algorithm}
\end{minipage}
\hfill
 \vspace{-1em}
\end{figure}

\subsection{Computational Algorithm}
\label{subsec:computation_algorithms}

From the dual problem \eqref{eq:dual_cont}, we can perform stochastic optimization as discussed in~\cite{genevay2016stochastic}. Nevertheless, SOT shows the most benefit in the discrete settings where SOT plans can be obtained efficiently. Therefore, we now focus our computational discussion to discrete case. We first start with deriving Sinkhorn algorithm for SROT by performing gradient-based optimization for $\mathbf{f}$ and $\mathbf{g}$ in~\eqref{eq:discrete_dual}.

\begin{proposition}
\label{prop:sr_ot_dual_updates}
The maximization of the duality~\eqref{eq:discrete_dual} over $(\mathbf f,\mathbf g)$ can be performed via coordinate ascent, where each block update admits the closed-form expressions
\begin{align}
\mathbf f_i
&=
\varepsilon \left[
\log \alpha_i
-
\log \sum_{j=1}^m
P^{\mathrm{SOT}}_{ij}
\exp\left(\frac{\mathbf g_j-C_{ij}}{\varepsilon}\right)
\right], \\
\mathbf g_j
&=
\varepsilon \left[
\log \beta_j
-
\log \sum_{i=1}^n
P^{\mathrm{SOT}}_{ij}
\exp\left(\frac{\mathbf f_i-C_{ij}}{\varepsilon}\right)
\right].
\end{align}
\end{proposition}
The proof of Proposition~\ref{prop:sr_ot_dual_updates} is given in Appendix~\ref{subsec:proof:prop:sr_ot_dual_updates}. Let $K = P^{\mathrm{SOT}} \odot \exp(-C/\varepsilon)$, where all exponential and logarithm operations are understood elementwise. Then the coordinate updates can be written as
$
\mathbf f
=
\varepsilon \left[
\log \boldsymbol{\alpha}
-
\log \left( K \exp(\mathbf g / \varepsilon) \right)
\right], \qquad 
\mathbf g
=
\varepsilon \left[
\log \boldsymbol{\beta}
-
\log \left( K^\top \exp(\mathbf f / \varepsilon) \right)
\right],
$
which is a log-stable Sinkhorn algorithm~\citep{benamou2015iterative}.  We would like to also discuss the Sinkhorn-style update via matrix scaling. Let $\mathbf u = \exp(\mathbf f / \varepsilon)$ and $\mathbf v = \exp(\mathbf g / \varepsilon)$, and
$K = P^{\mathrm{SOT}} \odot \exp(-C/\varepsilon)$, the dual updates are equivalent to the multiplicative scaling iterations
\begin{align}
\mathbf u
&=
\boldsymbol{\alpha} \oslash (K \mathbf v),  \qquad 
\mathbf v 
=
\boldsymbol{\beta} \oslash (K^\top \mathbf u),
\end{align}
where $\oslash$ denotes elementwise division. The transport plan admits the factorized form
\begin{align}
P_{\varepsilon,\mathrm{SOT}} = \operatorname{diag}(\mathbf u) K  \operatorname{diag}(\mathbf v),
\end{align}
and converges to a unique fixed point, as in EOT~\citep{sinkhorn1967diagonal}, due to the smoothing of $P^{\mathrm{SOT}}$. We summarize the algorithms for computing the SOT plan and the SROT plan in Algorithm~\ref{alg:SOT} and Algorithm~\ref{alg:SROT}, respectively. In practice, we run $T>0$ Sinkhorn iterations and apply early stopping when the maximum marginal violation falls below a prescribed tolerance. Since we recover the same scaling structure as in EOT, existing analyses of computational complexity directly apply~\citep{altschuler2017near}. {The only additional cost arises from computing the SOT plan: $\mathcal{O}(Ld(n+m))$ for the projections. In particular, we have $\mathcal{O}(L(n+m)\log(n+m))$ for sorting, and $\mathcal{O}(L(n+m))$ to accumulate the one-dimensional plans which have at most $n+m-1$ non-zero entries. This is negligible compared to the $\mathcal{O}(nm)$ cost of a single Sinkhorn iteration whenever $L \leq \min(n,m)$.}

{\textbf{Sparse Implementation.} When $\gamma=0$, the SOT plan is sparse by construction. For each direction $\theta_l$, the one-dimensional plan $P^{\mathrm{SOT}}_{\theta_l}$  has at most $n+m-1$ non-zero entries. Therefore, the plan $P^{\mathrm{SOT}}$ has at most $\min \{L(n+m-1),nm\}$ non-zero entries, which is a small fraction of $nm$ as soon as $L \ll \min(n,m)$. It is worth noting that $\min \{L(n+m-1),nm\}$ is a very loose upper bound, as one-dimensional plans can agree on a lot of matchings. The kernel $K = P^{\mathrm{SOT}} \odot \exp(-C/\varepsilon)$ in Algorithm~\ref{alg:SROT} vanishes outside the support of $P^{\mathrm{SOT}}$, and the updates $u \gets \boldsymbol{\alpha}/(Kv)$ and $v \gets \boldsymbol{\beta}/(K^\top u)$ only involve its non-zero entries. We can use a sparse matrix to store $P^{\mathrm{SOT}}$. We then run the Sinkhorn iterations with sparse matrix-vector products. One iteration then costs $\mathcal{O}(\min \{L(n+m),nm\})$ (again loose bound) in time and memory instead of $\mathcal{O}(nm)$.}

\subsection{Sliced-Regularized Optimal Transport Divergence}
\label{subsec:SROTdivergence}

We now discuss the transportation cost aspect of SROT. As in EOT, SROT provides an approximation of the optimal transport plan. Due to approximation error, the resulting transport cost is generally biased in the sense that it can be $0$ for two different probability measures. To address this issue, we introduce a debiased version, which we refer to as the SROT divergence.

\begin{definition}[SROT divergence] 
{Let $\mathcal{X}$ and $\mathcal{Y}$ be compact subsets of a metric space, $\mu \in \mathcal{P}(\mathcal{X})$ and $\nu \in \mathcal{P}(\mathcal{Y})$ be two probability measures, $c : (\mathcal{X}\cup\mathcal{Y}) \times (\mathcal{X}\cup\mathcal{Y}) \to \mathbb{R}_+$ a symmetric ground cost, $\varepsilon > 0$ a regularization parameter, and $\pi^{\mathrm{SOT}}$ a fixed reference plan between $(\mu,\nu)$. We define the SROT functional as follows:}
\begin{align}
\mathrm{OT}_{\varepsilon,\mathrm{SOT}}(\mu,\nu)
=
\inf_{\pi \in \Pi(\mu,\nu)}
\int_{\mathcal{X}\times\mathcal{Y}} c(x,y) \mathrm{d}\pi(x,y)
+
\varepsilon \mathrm{KL}\left(\pi \mid \pi^{\mathrm{SOT}}\right).
\end{align}
With the SROT functional, we define the SROT divergence as follows:
\begin{align}
\label{eq:SR_divergence}
\mathcal{S}_{\varepsilon,\mathrm{SOT}}(\mu,\nu)
=
\mathrm{OT}_{\varepsilon,\mathrm{SOT}}(\mu,\nu)
-\frac{1}{2}\mathrm{OT}_{\varepsilon,\mathrm{SOT}}(\mu,\mu)
-\frac{1}{2}\mathrm{OT}_{\varepsilon,\mathrm{SOT}}(\nu,\nu).
\end{align}
\end{definition}

\begin{figure}[!t]
    \centering
    \begin{tabular}{c}
       \includegraphics[width=1\linewidth]{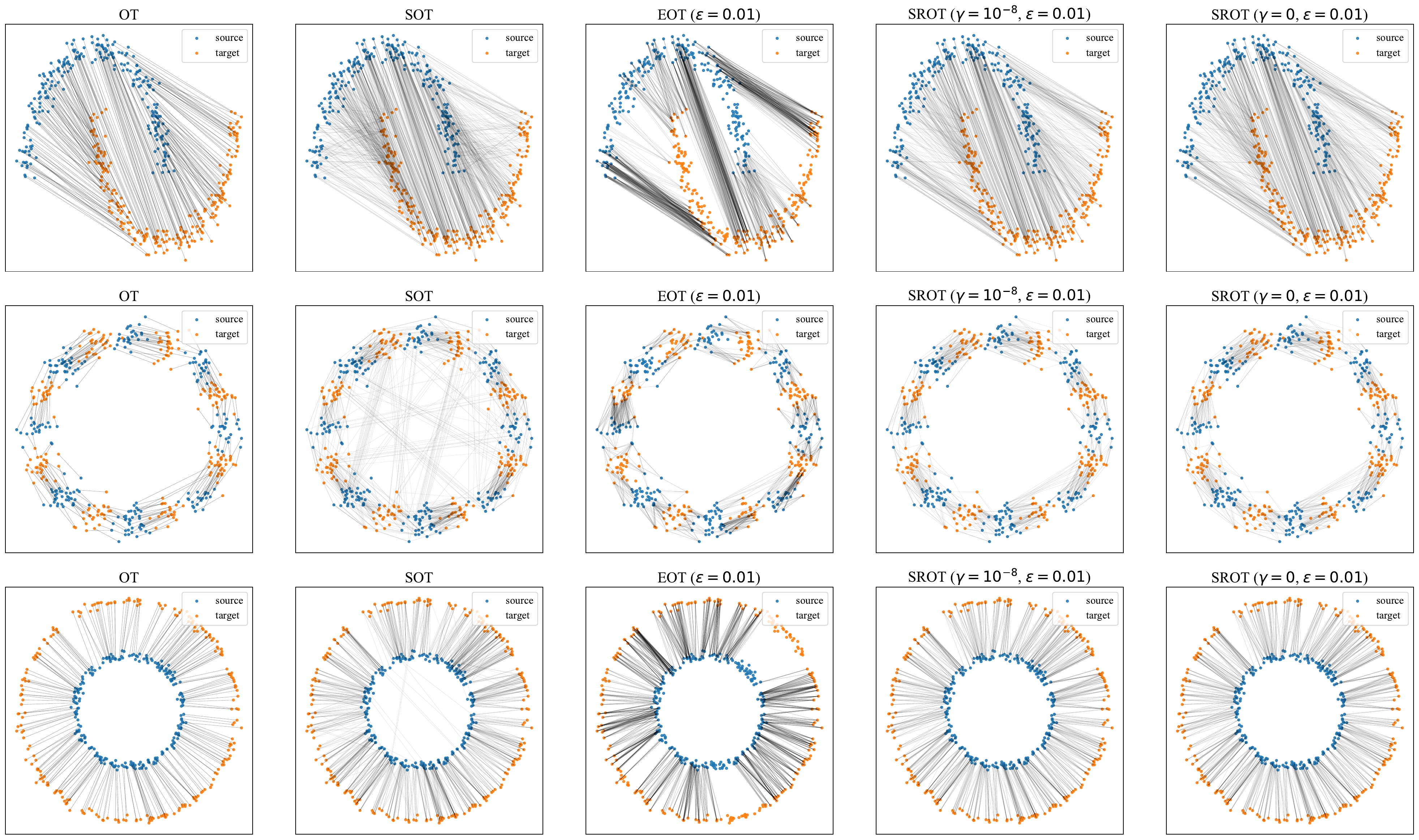}
    \end{tabular}
    \caption{{Visualization of transportation plans from OT, SOT, EOT, and SROT for synthetic datasets. For each plan only the $1{,}000$ largest entries are drawn, with opacity proportional to their mass.}}
    \label{fig:matching}
    \vspace{-1.5em}
\end{figure}
SROT is motivated by Sinkhorn divergence~\citep{genevay2018learning} from EOT. It contains the first term $\mathrm{OT}_{\varepsilon,\mathrm{SOT}}(\mu,\nu)$ for attraction and two terms $-\frac{1}{2}\mathrm{OT}_{\varepsilon,\mathrm{SOT}}(\mu,\mu)$ and $
-\frac{1}{2}\mathrm{OT}_{\varepsilon,\mathrm{SOT}}(\nu,\nu)$ for repulsion. The SROT divergence can also be seen as the interpolation of Wasserstein distance when $\varepsilon\to 0$ and a version of the maximum mean discrepancy (MMD)~\citep{gretton2012kernel} when $\varepsilon\to \infty$ (please see~\citep{feydy2019interpolating}).

{\begin{theorem}[Topological properties]
\label{theorem:topo}
Let $\mathcal{X}$ and $\mathcal{Y}$ be compact subsets of a metric space, $c : (\mathcal{X}\cup\mathcal{Y}) \times (\mathcal{X}\cup\mathcal{Y}) \to \mathbb{R}_+$ a symmetric Lipschitz cost, and let the fixed reference plan have a symmetric Lipschitz density $r : (\mathcal{X}\cup\mathcal{Y}) \times (\mathcal{X}\cup\mathcal{Y}) \to (0,\infty)$ with respect to the product measure, i.e., $\mathrm{d}\pi^{\mathrm{SOT}}(x,y) = r(x,y)\,\mathrm{d}\mu(x)\mathrm{d}\nu(y)$. Assume that
$k_\varepsilon(x,y) = r(x,y)\exp\!\big(-c(x,y)/\varepsilon\big)$ is a positive universal kernel on $\mathcal{X}\cup\mathcal{Y}$.
Then $\mathcal{S}_{\varepsilon,\mathrm{SOT}}$ is symmetric, non-negative, and satisfies:
$
\mu = \nu \quad \Longleftrightarrow \quad \mathcal{S}_{\varepsilon,\mathrm{SOT}}(\mu,\nu)=0,
$
for any $\varepsilon>0$. Moreover, it metrizes weak convergence:
$
\mu_n \rightharpoonup \mu
\quad \Longleftrightarrow \quad
\mathcal{S}_{\varepsilon,\mathrm{SOT}}(\mu_n,\mu) \to 0,
$
for any $\varepsilon>0$.
\end{theorem}}
{The proof of Theorem~\ref{theorem:topo} is given in Appendix~\ref{subsec:proof:theorem:topo}: with the reference fixed, $\mathcal{S}_{\varepsilon,\mathrm{SOT}}$ is the Sinkhorn divergence of the adjusted cost $c - \varepsilon\log r$, so the result follows from Theorem~1 of~\cite{feydy2019interpolating}. The statement is for a fixed reference i.e., it does not cover a reference recomputed along the sequence $(\mu_n)$, since the plug-in SOT plan is not weakly continuous in its arguments. Theorem~\ref{theorem:topo} guarantees the usage of SROT divergence as a loss for estimating parameters.}

\section{Experiments}
\label{sec:experiments}
In this section, we aim to compare SROT with EOT in approximating OT.  In particular, we focus on the Euclidean setting where $c(x,y)=\|x-y\|_2$ as it appears widely in practice. We conduct comparison on synthetic datasets in Section~\ref{subsec:synthetic_data} and color transfer in Section~\ref{subsec:color_transfer}. In addition, we compare SROT divergence with  Sinkhorn divergence in gradient flow in Section~\ref{subsec:gradient_flow}. Additional experiments mentioned in the main paper is given in Appendix~\ref{sec:additional_experiments}.  All experiments are conducted on a  HP Omen 25L desktop.

\subsection{Synthetic Data}
\label{subsec:synthetic_data}
We construct discrete measures $\mu$ and $\nu$ on $\mathbb{R}^d$ by sampling $n$ points from three synthetic datasets with distinct geometries: \emph{half moon}, consisting of two noisy, interleaved half-moons defining $\mu$ and $\nu$; \emph{8 Gaussians}, where each distribution is a mixture of eight isotropic Gaussian components; and \emph{two rings}, where $\mu$ and $\nu$ are supported on concentric circles with different radii and small radial perturbations. For SOT, we adopt a uniform slicing distribution by default. Empirically, SROT with this choice performs robustly compared to alternative slicing strategies, even though some may provide a closer reference to the exact OT plan. We refer the reader to the ablation study in Figure~\ref{fig:L} (Appendix~\ref{sec:additional_experiments}) for further details. We set the number of projections to $L = 100$ for SOT.

\begin{figure}[!t]
    \centering
    \begin{tabular}{c}
       \includegraphics[width=1\linewidth]{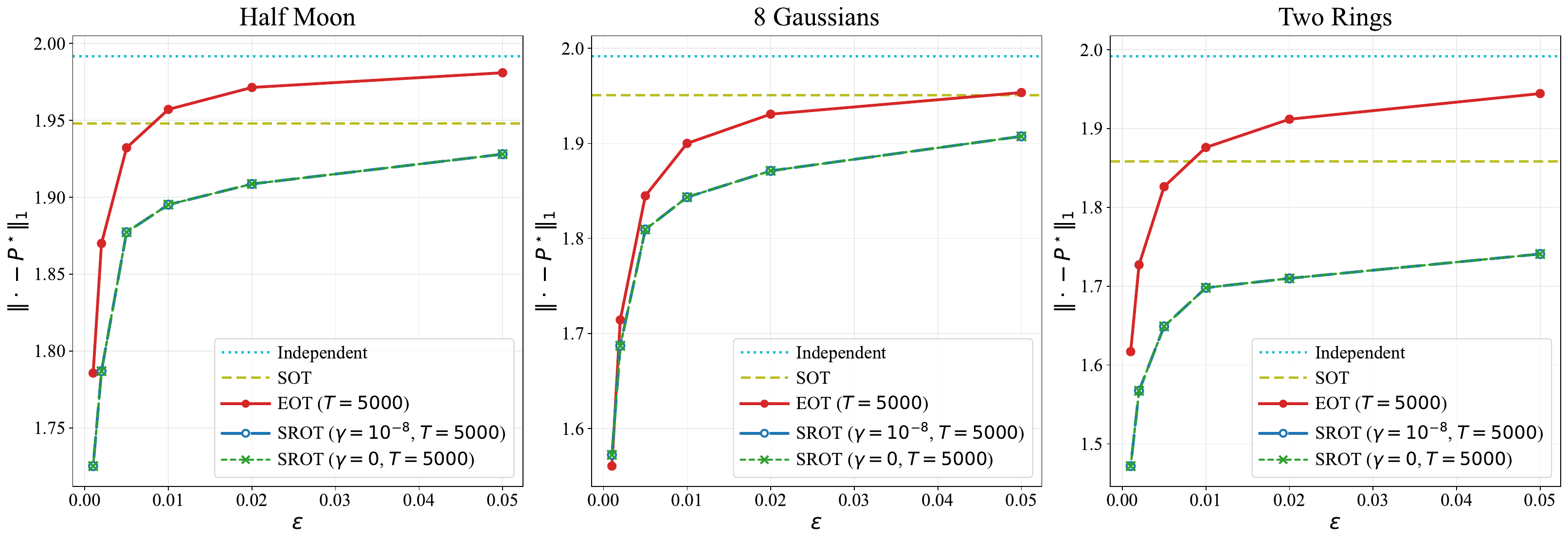} \\
          \includegraphics[width=1\linewidth]{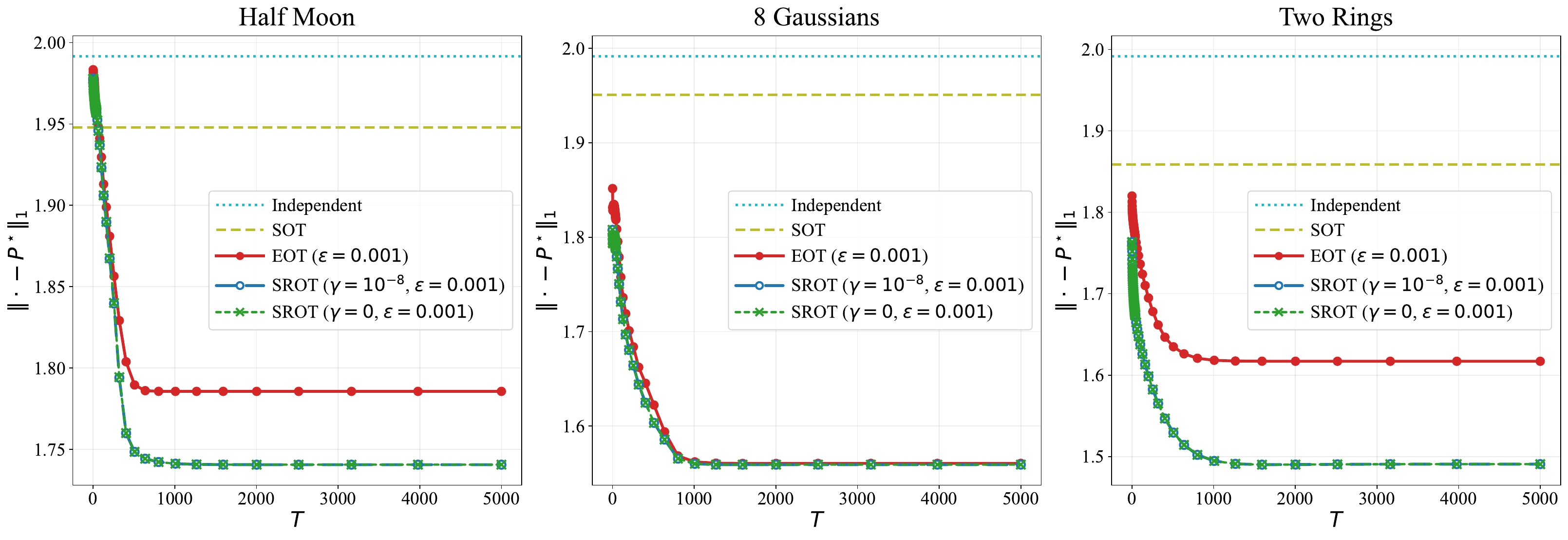} \\
          
    \end{tabular}
       \vspace{-1em}
    \caption{Ablation studies of varying the regularization strengths ($\varepsilon$) and  Sinkhorn iterations ($T$)}
    \label{fig:eps_T}
       \vspace{-1em}
\end{figure}

\begin{figure}[!t]
    \centering
    \begin{tabular}{c}
       \includegraphics[width=1\linewidth]{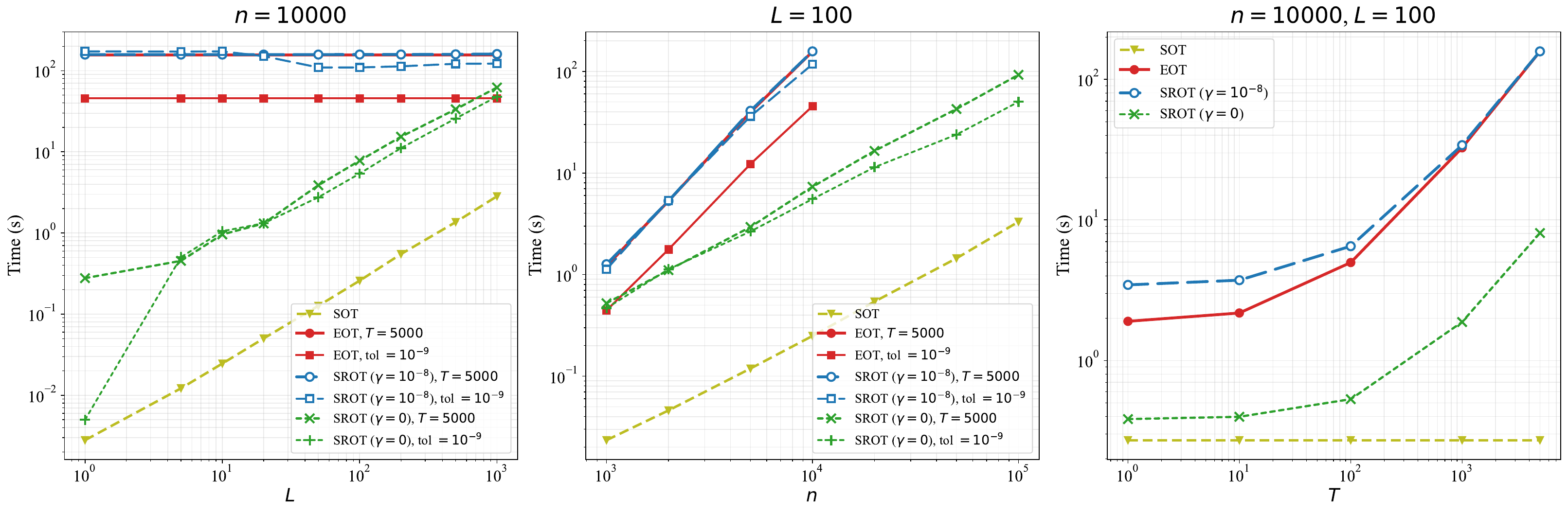} 
          
    \end{tabular}
       \vspace{-1em}
    \caption{Computational speed on half moon ($\varepsilon = 0.001$) when varying the number of projections $L$ ($n = 10^4$), the number of atoms $n$ ($L = 100$), and the number of iterations $T$ ($n = 10^4$, $L = 100$).}
    \label{fig:speed}
       \vspace{-1em}
\end{figure}

\textbf{Visualization of transportation plans.} We visualize the transportation plans obtained from exact OT, SOT, EOT, and SROT in Figure~\ref{fig:matching}. For EOT and SROT, we report results with $\varepsilon =  0.01$ {(results for $\varepsilon =  0.001$ are in Figure~\ref{fig:matching_0001} in Appendix~\ref{sec:additional_experiments})} and $T = 5000$ Sinkhorn iterations. The results show that SROT yields visually more accurate transport plans than EOT, particularly for larger values of $\varepsilon$. Moreover, SROT remains effective even when the SOT approximation is relatively poor, as illustrated in the 8-Gaussians example. {We observe that SROT $(\gamma=1e-8)$ and SROT $(\gamma=0)$ give nearly identical plans.}

\textbf{Regularization strength ($\varepsilon$).}
We fix the number of Sinkhorn iterations to $T = 5000$ and vary the entropic regularization parameter $\varepsilon$. For each $\varepsilon$, we compare EOT and SROT using the approximation error $\|\cdot - P^\star\|_1$, where $P^\star$ denotes the exact OT plan. We also include two horizontal baselines under the same metric: the independent product coupling and the SOT plan. The results are shown in the first row of Figure~\ref{fig:eps_T}. We observe that the SOT plan is consistently closer to the OT plan than the independent coupling. Consequently, SROT outperforms EOT across nearly all choices of $\varepsilon$, with the exception of a single setting at very small $\varepsilon$ in the 8 Gaussians case. As $\varepsilon$ increases, EOT gradually approaches the independent coupling, while SROT converges toward the SOT plan.  {We again observe that SROT $(\gamma=1e-8)$ and SROT $(\gamma=0)$ give nearly identical plans.}

\textbf{Sinkhorn iterations ($T$).}
We fix $\varepsilon = 0.001$ and record the approximation error to the true OT plan as a function of the number of Sinkhorn iterations $T$, again comparing EOT and SROT. The results are shown in the second row of Figure~\ref{fig:eps_T}. Overall, SROT achieves a lower approximation error than EOT at convergence, while both methods exhibit comparable convergence rates.


{\textbf{Computational Speed.} We report the wall-clock runtime of SROT and EOT in Figure~\ref{fig:speed}. The cost of computing the SOT plan is negligible: at $n = 10^4$ it is below the cost of forming the kernel of EOT. With dense Sinkhorn iterations, SROT and EOT have the same cost per iteration, and SROT is slower than EOT at $\varepsilon = 0.001$. With the sparse iterations ($\gamma = 0$), an iteration is much cheaper than a dense one at $n = 10^4$, so SROT reaches the tolerance faster than EOT for $L \leq 500$. Moreover, it runs at $n = 10^5$ where the dense kernel of EOT does not fit in memory. It is  worth nothing that we do not use parallel computation for SOT here while it is applicable i.e., projections can be computed independently. }
\begin{table}[!t]
\centering
\caption{$\mathbb{L}_1$ error, relative transportation cost gap and wall-clock time
versus exact OT (mean over 3 pairs of images) in color transfer with $10{,}000$
colors ($L=100$, $T=5000$).}
\label{tab:color_10k}
\scalebox{0.8}{
\begin{tabular}{l ccc ccc ccc}
\toprule
 & \multicolumn{3}{c}{$\epsilon = 1$} & \multicolumn{3}{c}{$\epsilon = 0.1$} & \multicolumn{3}{c}{$\epsilon = 0.01$} \\
\cmidrule(lr){2-4} \cmidrule(lr){5-7} \cmidrule(lr){8-10}
Method & $\mathbb{L}_1$ & gap & time (s) & $\mathbb{L}_1$ & gap & time (s) & $\mathbb{L}_1$ & gap & time (s) \\
\midrule
SOT & \multicolumn{9}{c}{$\mathbb{L}_1 = 1.998$, \; gap $= +61.3\%$, \; time $= 0.26$ s} \\
\midrule
EOT & 2.000 & $+139.2\%$ & 1.88 & 1.999 & $+43.5\%$ & 3.30 & 1.994 & $+7.4\%$ & 10.34 \\
SROT ($\gamma=10^{-8}$) & 1.997 & $\mathbf{+49.2\%}$ & 2.48 & 1.996 & $\mathbf{+21.7\%}$ & 4.57 & 1.989 & $\mathbf{+5.4\%}$ & 18.49 \\
SROT ($\gamma=0$) & \textbf{1.997} & $+49.3\%$ & \textbf{0.43} & \textbf{1.996} & $+21.9\%$ & \textbf{0.52} & \textbf{1.989} & $\mathbf{+5.4\%}$ & \textbf{1.27} \\
\bottomrule
\end{tabular}}
\end{table}

\begin{figure}[!t]
    \centering
    \setlength{\tabcolsep}{0pt}
    \renewcommand{\arraystretch}{0}
    \newcommand{\imw}{0.2\textwidth}
    \scalebox{1}{
    \begin{tabular}{c}
    \includegraphics[width=1\linewidth]{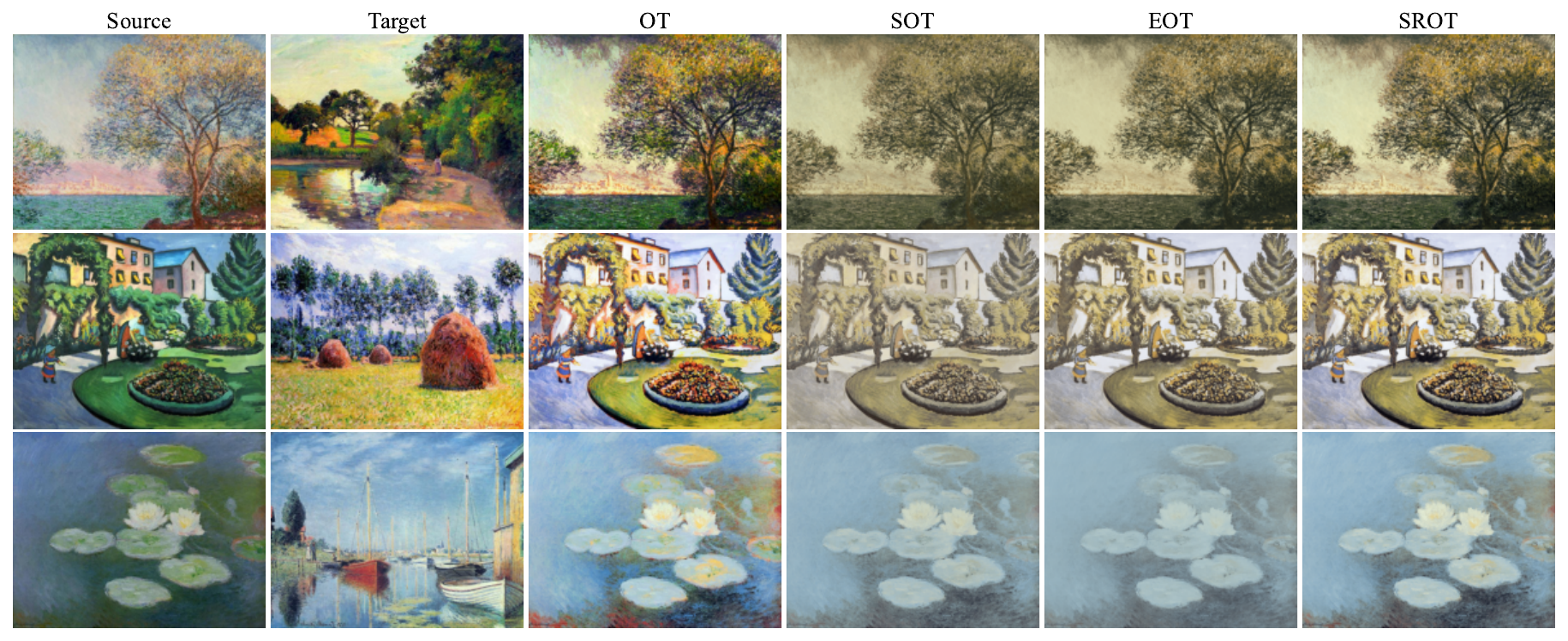}
    \end{tabular}
}
    \caption{\footnotesize{Color transfer results of OT, SOT, EOT, and SROT.}}
    \label{fig:color_transfer}
    \vspace{-2em}
\end{figure}
\subsection{Color Transfer}
\label{subsec:color_transfer}
{The color transfer task is modeled as an OT problem via the point cloud
representation of each image in the normalized RGB color space $[0,1]^3$. We sample
$10{,}000$ pixels uniformly at random with the equal weights from each image, hence
each couple of images is the $10{,}000 \times 10{,}000$ transport problem. We
compare SOT, EOT, and SROT with $\gamma = 10^{-8}$ and $\gamma = 0$ using the
entropic regularization parameter $\varepsilon \in \{1, 10^{-1}, 10^{-2}\}$,
$T = 5000$ Sinkhorn iterations, and $10^{-9}$ marginal tolerance. The color
transfer result is found through barycentric projection: each source pixel is mapped
to the nearest sampled color which then is recolored according to the row-normalized
transport plan image. Since the $\mathbb{L}_1$ error is almost saturated to its
maximum value of 2 at this size, we measure and report the relative difference
between the transportation cost of the approximation and the optimal plan, along with
the computation time averaged over three image couples (Table
\ref{tab:color_10k}). Transferred images at $\varepsilon = 0.1$ can be found in
Figure \ref{fig:color_transfer}.  SROT reduces the cost gap of EOT at every
$\varepsilon$, by about half for $\varepsilon \geq 0.1$.  SROT with $\gamma = 0$ is the fastest method, while the
SOT plan alone washes the colors out.}

\vspace{-1em}

\subsection{Gradient Flow}
\label{subsec:gradient_flow}

\begin{figure}[!t]
    \centering
    \setlength{\tabcolsep}{0pt}
    \renewcommand{\arraystretch}{0}
    \newcommand{\imw}{0.2\textwidth}
    \scalebox{0.95}{
    \begin{tabular}{c}   \includegraphics[width=0.95\linewidth]{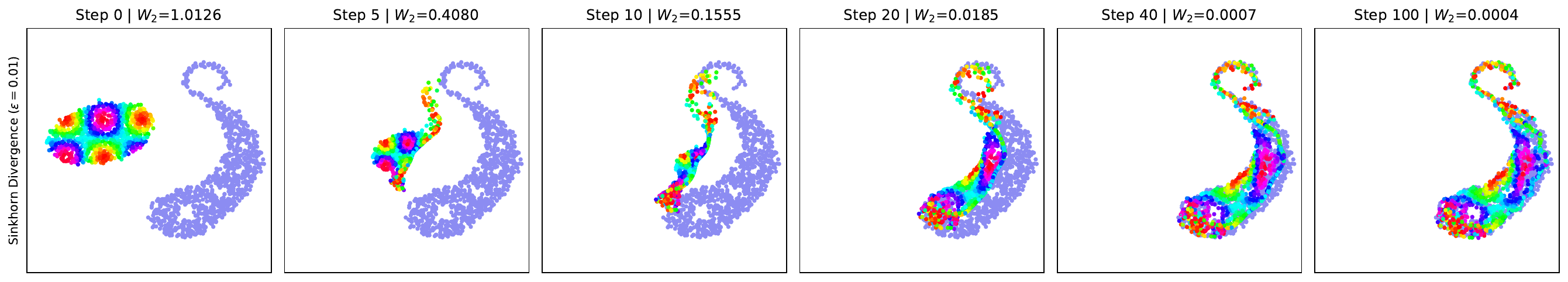} \\
\includegraphics[width=0.95\linewidth]{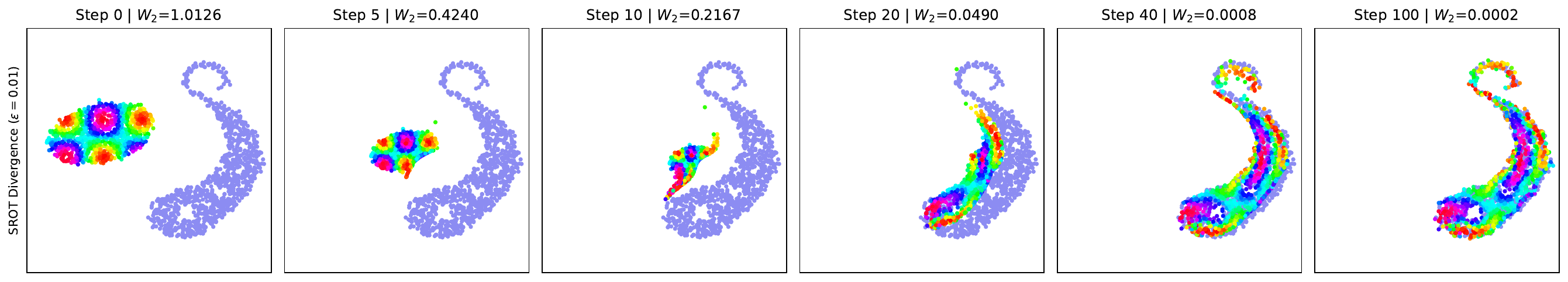} \\
\includegraphics[width=0.95\linewidth]{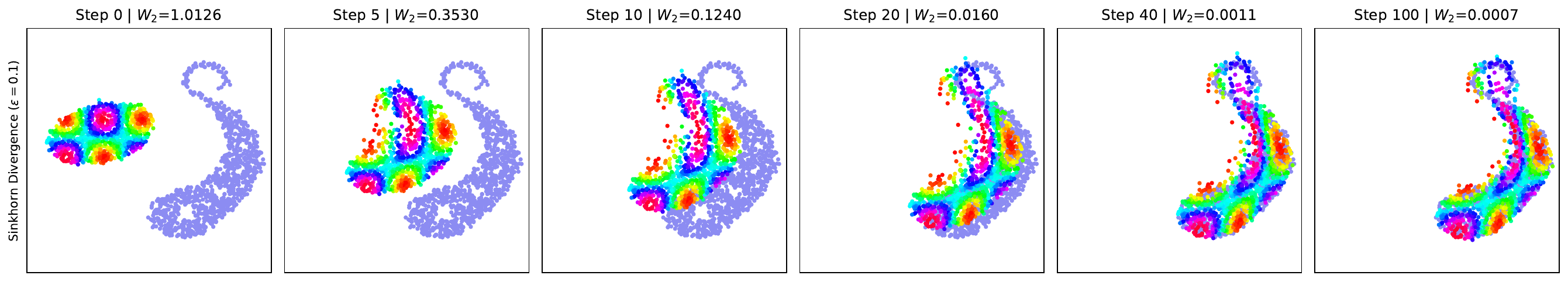} \\
\includegraphics[width=0.95\linewidth]{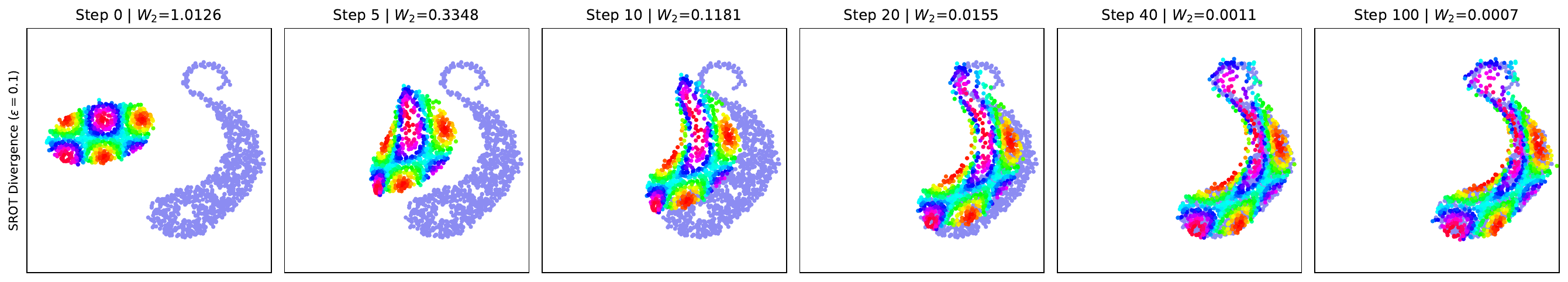} 
        
    \end{tabular}
}\vspace{-0.5em}
    \caption{\footnotesize{Gradient flows of Sinkhorn divergence and SROT divergence.}}
    \label{fig:gd}
    \vspace{-2em}
\end{figure}

We follow the gradient-flow setup of~\cite{feydy2019interpolating,santambrogio2015optimal} and consider the evolution of an empirical measure $\mu(t)$ toward a fixed target measure $\nu$ by minimizing a discrepancy functional $\mathcal{D}(\mu(t),\nu)$. In our implementation, both measures are represented by point clouds in $\mathbb{R}^2$ with equal masses:
$
\nu=\frac{1}{n}\sum_{j=1}^{n}\delta_{Y_j},
\mu(t)=\frac{1}{n}\sum_{i=1}^{n}\delta_{X_i(t)},
$
where $n=1000$. Starting from $X(0)$, we integrate
$
\dot X(t)=-n\nabla_{X(t)}\left[\mathcal{D}\left(\frac{1}{n}\sum_{i=1}^{n}\delta_{X_i(t)},\nu\right)\right]
$
with an explicit Euler scheme:
$
X^{k+1}=X^k-\eta\nabla_X\mathcal{D}(X^k,Y), \eta=0.05.
$
We run the flow for $100$ iterations. We evaluate both the  Sinkhorn divergence and  SROT divergence using Wasserstein distance.  We report the result for $\varepsilon\in [0.01, 0.1]$ in Figure~\ref{fig:gd} and for $\varepsilon =1$ in Figure~\ref{fig:gd_appendix} in Appendix~\ref{sec:additional_experiments}.  We observe that SROT divergence provides weaker repulsion than  Sinkhorn divergence especially for small $\varepsilon$ e.g., 0.01. The reason is that SOT plan  is  not uniform.  For $\varepsilon\in (0.1,1)$, SROT divergence makes the flow converge faster in the sense of Wasserstein distance,

\vspace{-1em}
\section{Conclusion}
\label{sec:conclusion}

\vspace{-0.5em}
We propose sliced-regularized optimal transport (SROT), a framework that leverages a smoothed SOT plan as an informative prior to improve regularized OT. SROT retains the computational efficiency of EOT via a Sinkhorn-style algorithm while providing more accurate approximations to the true OT plan. We establish its theoretical properties and introduce the SROT divergence. Experiments show that SROT is consistently better than  both EOT and SOT across tasks such as synthetic matching, color transfer, and gradient flows. A limitation of SROT is its reliance on the quality of the prior SOT plan, which is still an active area of research, especially in high-dimensional settings. Future work includes studying statistical properties of SROT and its divergence (e.g., sample complexity and central limit behavior), and handling high-dimensionality.

\subsection*{AI use statement}

In this work, we used generative AI tools for editing the paper and for coding. We take responsibility for the final content of this work,
including text, claims, or artifacts produced with the aid of generative AI.

\appendix

\section{Proofs}
\subsection{Proof of Theorem~\ref{theorem:dual_cont}}
\label{subsec:proof:theorem:dual_cont}
We introduce Lagrange multipliers $f \in \mathcal{C}(\mathcal{X})$ and $g \in \mathcal{C}(\mathcal{Y})$ for the marginal constraints. The Lagrangian is
\begin{align}
\mathcal{L}(\pi,f,g)
&= \int_{\mathcal{X}\times \mathcal{Y}} \bigl[c(x,y) - f(x) - g(y)\bigr] \mathrm{d}\pi(x,y)
\nonumber \\
&+ \varepsilon \int_{\mathcal{X}\times \mathcal{Y}}  \log\left(\frac{\mathrm{d}\pi}{\mathrm{d}\pi^{\mathrm{SOT}}}(x,y)\right) \mathrm{d}\pi (x,y)
+ \int_\mathcal{X} f(x) \mathrm{d}\mu(x) + \int_\mathcal{Y} g(y) \mathrm{d}\nu(y).
\end{align}
Let $\rho = \frac{\mathrm{d}\pi}{\mathrm{d}\pi^{\mathrm{SOT}}}$, we have
\begin{align}
\mathcal{L}(\pi,f,g)
&= \int_{\mathcal{X}\times \mathcal{Y}} \Big[
(c(x,y) - f(x) - g(y))\rho(x,y) + \varepsilon \rho(x,y) \log \rho(x,y)
\Big] \mathrm{d}\pi^{\mathrm{SOT}}(x,y) \nonumber\\
&+ \int_{\mathcal{X}} f(x) \mathrm{d}\mu(x) + \int_{\mathcal{Y}} g(y) \mathrm{d}\nu(y).
\end{align}
Let $h(x,y) = c(x,y) - f(x) - g(y)$ and we minimize pointwise over $\rho$:
\begin{align}
\inf_{\rho \ge 0}
\left[
h\rho + \varepsilon \rho \log \rho
\right].
\end{align}
The first-order optimality condition is
\begin{align}
h + \varepsilon(\log\rho + 1) = 0,
\end{align}
which yields
\begin{align}
\rho^*(x,y)
= \exp\left(-\frac{h(x,y)}{\varepsilon} - 1\right).
\end{align}
Substituting $\rho^*$ into the objective gives
\begin{align}
h\rho^* + \varepsilon \rho^* \log \rho^*
= -\varepsilon \rho^*
= -\varepsilon \exp\left(-\frac{h(x,y)}{\varepsilon} - 1\right).
\end{align}
Absorbing the constant factor $e^{-1}$ into the dual potentials yields
\begin{align}
\inf_{\rho \ge 0}
\left[
h\rho + \varepsilon \rho \log \rho
\right]
= -\varepsilon \exp\left(-\frac{h(x,y)}{\varepsilon}\right).
\end{align}
Therefore,
\begin{align}
\inf_{\pi \ge 0} \mathcal{L}(\pi,f,g)
&=
\int_{\mathcal{X}} f(x) \mathrm{d}\mu(x)
+ \int_{\mathcal{Y}} g(y) \mathrm{d}\nu(y) \nonumber\\&
- \varepsilon \int_{\mathcal{X}\times\mathcal{Y}}
\exp\left(\frac{f(x)+g(y)-c(x,y)}{\varepsilon}\right)
\mathrm{d}\pi^{\mathrm{SOT}}(x,y).
\end{align}

Maximizing over $(f,g)$ yields the dual problem. Strong duality follows from convexity of the primal objective and the existence of a feasible $\pi \ll \pi^{\mathrm{SOT}}$. The optimal coupling satisfies
\begin{align}
\frac{\mathrm{d}\pi^\star}{\mathrm{d}\pi^{\mathrm{SOT}}}(x,y)
=
\exp\left(\frac{f^*(x)+g^*(y)-c(x,y)}{\varepsilon}\right),
\end{align}
which completes the proof.

\subsection{Proof of Proposition~\ref{prop:sr_ot_dual_updates}}
\label{subsec:proof:prop:sr_ot_dual_updates}

Consider the dual objective
\begin{align}
\mathcal{L}(\mathbf f,\mathbf g)
=
\langle \mathbf f,\boldsymbol{\alpha}\rangle
+ \langle \mathbf g,\boldsymbol{\beta}\rangle
- \varepsilon \sum_{i=1}^n \sum_{j=1}^m
P^{\mathrm{SOT}}_{ij}
\exp\left(\frac{f_i+g_j - C_{ij}}{\varepsilon}\right).
\end{align}

Taking derivatives with respect to $f_i$ and $g_j$ and setting them to $0$ yields
\begin{align}
\frac{\partial \mathcal{L}}{\partial f_i}
&=
\alpha_i
-
\sum_{j=1}^m
P^{\mathrm{SOT}}_{ij}
\exp\left(\frac{f_i+g_j-C_{ij}}{\varepsilon}\right)
=0, \\
\frac{\partial \mathcal{L}}{\partial g_j}
&=
\beta_j
-
\sum_{i=1}^n
P^{\mathrm{SOT}}_{ij}
\exp\left(\frac{f_i+g_j-C_{ij}}{\varepsilon}\right)
=0.
\end{align}
Rearranging the first equation, we have
\begin{align}
\alpha_i
&=
\sum_{j=1}^m
P^{\mathrm{SOT}}_{ij}
\exp\left(\frac{f_i+g_j-C_{ij}}{\varepsilon}\right) =
e^{f_i/\varepsilon}
\sum_{j=1}^m
P^{\mathrm{SOT}}_{ij}
\exp\left(\frac{g_j-C_{ij}}{\varepsilon}\right),
\end{align}
which leads to
\begin{align}
f_i
=
\varepsilon \left[
\log \alpha_i
-
\log \sum_{j=1}^m
P^{\mathrm{SOT}}_{ij}
\exp\left(\frac{g_j-C_{ij}}{\varepsilon}\right)
\right].
\end{align}
Similarly, rearranging the first equation, we have
\begin{align}
g_j
=
\varepsilon \left[
\log \beta_j
-
\log \sum_{i=1}^n
P^{\mathrm{SOT}}_{ij}
\exp\left(\frac{f_i-C_{ij}}{\varepsilon}\right)
\right].
\end{align}
We conclude the proof.

\subsection{Proof of Theorem~\ref{theorem:topo}}
\label{subsec:proof:theorem:topo}
{We recall that the reference plan is fixed and has density $r$ with respect to the product measure. Let $\mathcal{Z} = \mathcal{X}\cup\mathcal{Y}$, which is compact, and let $c_r(x,y) = c(x,y) - \varepsilon \log r(x,y)$ be the adjusted ground cost on $\mathcal{Z}\times\mathcal{Z}$. For $(\xi,\eta) \in \{(\mu,\nu),(\mu,\mu),(\nu,\nu)\}$ and $\pi \in \Pi(\xi,\eta)$ with $\pi \ll \xi \otimes \eta$, we have
\begin{align}
\mathrm{KL}\left(\pi \mid r\cdot(\xi\otimes\eta)\right)
= \int_{\mathcal{Z}\times\mathcal{Z}} \log\left(\frac{\mathrm{d}\pi}{\mathrm{d}(\xi\otimes\eta)}(x,y)\right) \mathrm{d}\pi(x,y) - \int_{\mathcal{Z}\times\mathcal{Z}} \log r(x,y)\, \mathrm{d}\pi(x,y),
\end{align}
which yields
\begin{align}
\int_{\mathcal{Z}\times\mathcal{Z}} c(x,y)\,\mathrm{d}\pi(x,y) + \varepsilon\,\mathrm{KL}\left(\pi \mid r\cdot(\xi\otimes\eta)\right)
= \int_{\mathcal{Z}\times\mathcal{Z}} c_r(x,y)\,\mathrm{d}\pi(x,y) + \varepsilon\,\mathrm{KL}\left(\pi \mid \xi\otimes\eta\right).
\end{align}
Taking the infimum over $\pi \in \Pi(\xi,\eta)$, we obtain that $\mathrm{OT}_{\varepsilon,\mathrm{SOT}}(\xi,\eta)$ is the EOT functional in~\eqref{eq:EOT} with ground cost $c_r$. Therefore, $\mathcal{S}_{\varepsilon,\mathrm{SOT}}$ is the Sinkhorn divergence with ground cost $c_r$ on $\mathcal{P}(\mathcal{Z})$. Since $\mathcal{Z}$ is compact and $r$ is continuous and positive, $r$ is bounded away from $0$, hence $\log r$ is Lipschitz and $c_r$ is a symmetric Lipschitz cost. Moreover, the Gibbs kernel of $c_r$ is
\begin{align}
\exp\left(-\frac{c_r(x,y)}{\varepsilon}\right) = r(x,y)\exp\left(-\frac{c(x,y)}{\varepsilon}\right) = k_\varepsilon(x,y),
\end{align}
which is a positive universal kernel on $\mathcal{Z}$ by assumption. Applying Theorem~1 in~\cite{feydy2019interpolating} with the ground cost $c_r$, we obtain that $\mathcal{S}_{\varepsilon,\mathrm{SOT}}$ is symmetric, non-negative, satisfies $\mathcal{S}_{\varepsilon,\mathrm{SOT}}(\mu,\nu) = 0 \Leftrightarrow \mu = \nu$, and metrizes weak convergence. We conclude the proof.}

{\subsection{Proof of Proposition~\ref{prop:approx}}
\label{subsec:proof:prop:approx}
From Theorem~\ref{theorem:dual_cont} in the discrete case, the regularized plan has the form
\begin{align}
P_{\varepsilon,ij} = P_{0,ij} \exp\left(\frac{\mathbf{f}_i + \mathbf{g}_j - C_{ij}}{\varepsilon}\right),
\end{align}
where $(\mathbf{f},\mathbf{g})$ are optimal dual potentials. Let $P \in \Gamma(\boldsymbol{\alpha},\boldsymbol{\beta})$ be any feasible plan, we have
\begin{align}
\mathrm{KL}(P \mid P_0) - \mathrm{KL}(P \mid P_\varepsilon)
= \sum_{i=1}^n \sum_{j=1}^m P_{ij} \log\left(\frac{P_{\varepsilon,ij}}{P_{0,ij}}\right)
= \frac{\langle \mathbf{f},\boldsymbol{\alpha}\rangle + \langle \mathbf{g},\boldsymbol{\beta}\rangle - \langle C, P\rangle}{\varepsilon},
\end{align}
where we use $P\mathbf{1} = \boldsymbol{\alpha}$ and $P^\top \mathbf{1} = \boldsymbol{\beta}$. Taking $P = P_\varepsilon$ in the above equation, we have
\begin{align}
\mathrm{KL}(P_\varepsilon \mid P_0) = \frac{\langle \mathbf{f},\boldsymbol{\alpha}\rangle + \langle \mathbf{g},\boldsymbol{\beta}\rangle - \langle C, P_\varepsilon\rangle}{\varepsilon}.
\end{align}
Subtracting the two equations and rearranging, we obtain for any $P \in \Gamma(\boldsymbol{\alpha},\boldsymbol{\beta})$:
\begin{align}
\label{eq:three_point}
\langle C, P\rangle + \varepsilon\, \mathrm{KL}(P \mid P_0)
= \langle C, P_\varepsilon\rangle + \varepsilon\, \mathrm{KL}(P_\varepsilon \mid P_0) + \varepsilon\, \mathrm{KL}(P \mid P_\varepsilon).
\end{align}
We now take $P = P^\star$ in~\eqref{eq:three_point} and drop the two non-negative KL terms on the right-hand side, which gives
\begin{align}
\langle C, P_\varepsilon\rangle - \langle C, P^\star\rangle
= \varepsilon\, \mathrm{KL}(P^\star \mid P_0) - \varepsilon\, \mathrm{KL}(P_\varepsilon \mid P_0) - \varepsilon\, \mathrm{KL}(P^\star \mid P_\varepsilon)
\leq \varepsilon\, \mathrm{KL}(P^\star \mid P_0).
\end{align}
We conclude the proof.
}

\section{Additional Experiments}
\label{sec:additional_experiments}

\begin{figure}[!t]
    \centering
    \begin{tabular}{c}
       \includegraphics[width=1\linewidth]{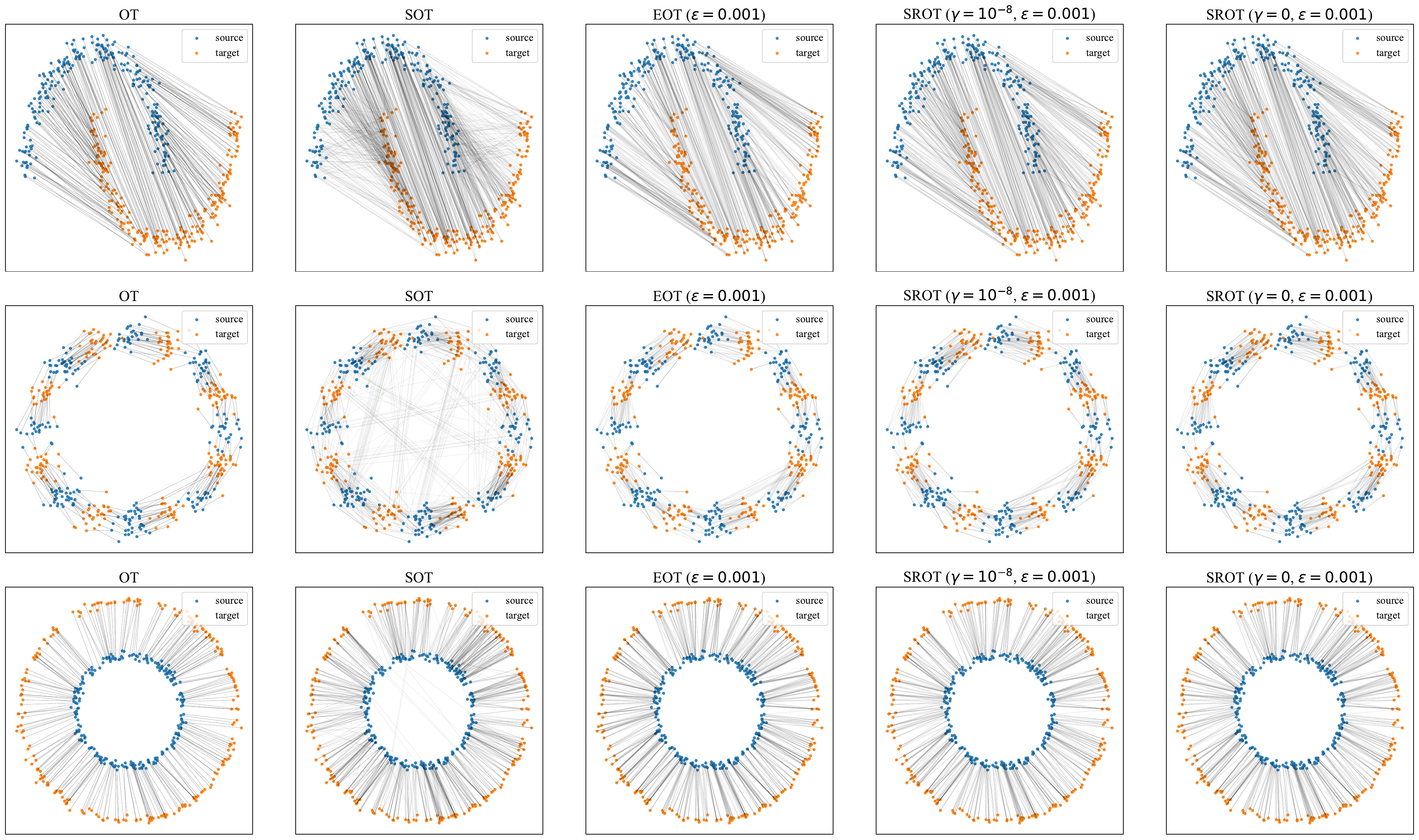}
    \end{tabular}
    \caption{{Visualization of transportation plans from OT, SOT, EOT, and SROT for synthetic datasets. For each plan only the $1{,}000$ largest entries are drawn, with opacity proportional to their mass.}}
    \label{fig:matching_0001}
    \vspace{-1em}
\end{figure}

\begin{figure}[!t]
    \centering
    \begin{tabular}{c}
          \includegraphics[width=1\linewidth]{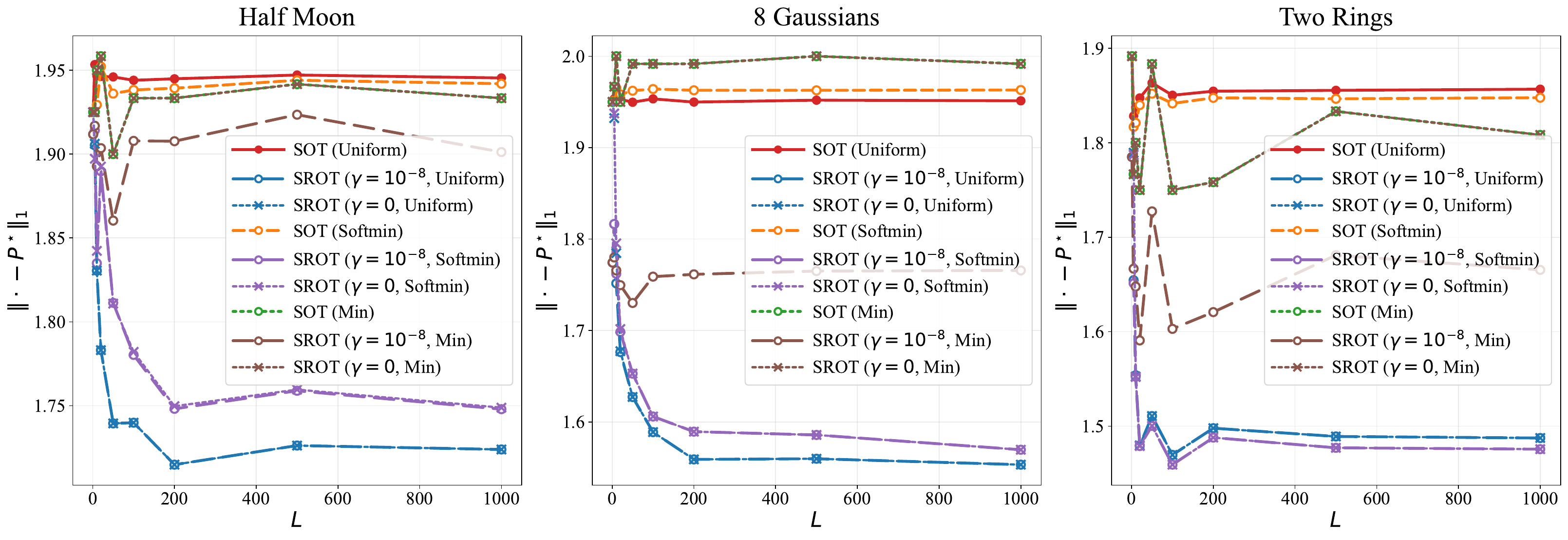}
    \end{tabular}
    \caption{Ablation study of varying the number of projections $L$.}
    \label{fig:L}
\end{figure}

\textbf{Number of projections ($L$) and SOT reference plan.}
We study how the number of projections $L$ affects plan quality before and after Sinkhorn for SROT, with $(\varepsilon, T) = (0.001, 5000)$ fixed. For each $L$, a single random draw of $L$ unit directions defines three reference couplings on the same slices: uniform averaging (SOT (Uniform)), a softmin variant (SOT (Softmin)), and the one-dimensional plan with minimum slice cost (SOT (Min)). For each reference, we report $\|\cdot - P^\star\|_1$ for both the initialization and the SROT solution. Results are shown in Figure~\ref{fig:L}. We observe that, at initialization, uniform averaging does not necessarily best approximate the true OT plan. Nevertheless, after Sinkhorn, SROT initialized with the uniform SOT is typically the most accurate, except in the two-rings case. This may be due to the greater smoothness of the corresponding SOT plan. For both uniform and softmin initializations, SROT improves as $L$ increases, even when the initial plans do not. Overall, we recommend uniform SOT as a default choice, while noting that it may not be optimal and merits further exploration.

\begin{figure}[!t]
    \centering
    \begin{tabular}{c}
          \includegraphics[width=1\linewidth]{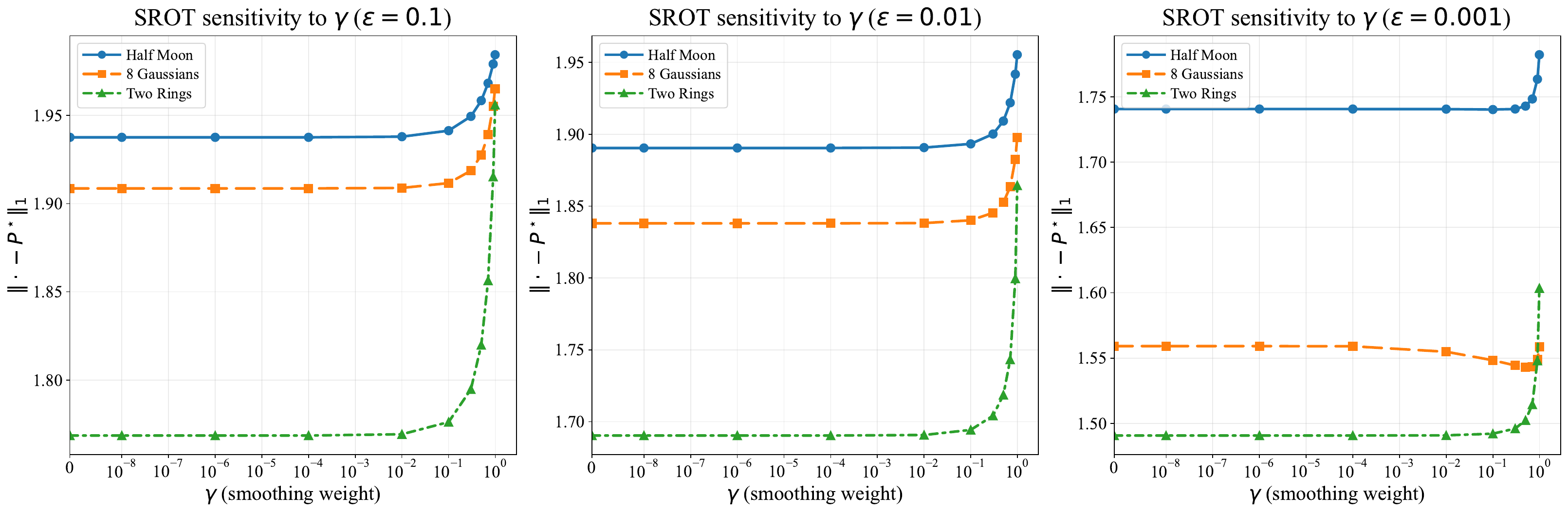}
    \end{tabular}
    \caption{Sensitivity of SROT to the smoothing weight $\gamma$ of the reference plan.}
    \label{fig:gamma}
\end{figure}
{
\textbf{Smoothing Sensitivity.} In practice, we find that $\gamma = 0$ works well in approximating OT, and the plan is unchanged for any $\gamma$ up to about $10^{-2}$ (see Figure~\ref{fig:gamma}).
}



\begin{figure}[!t]
    \centering
    \setlength{\tabcolsep}{0pt}
    \renewcommand{\arraystretch}{0}
    \newcommand{\imw}{0.2\textwidth}
    \scalebox{1}{
    \begin{tabular}{c}
     
    \includegraphics[width=1\linewidth]{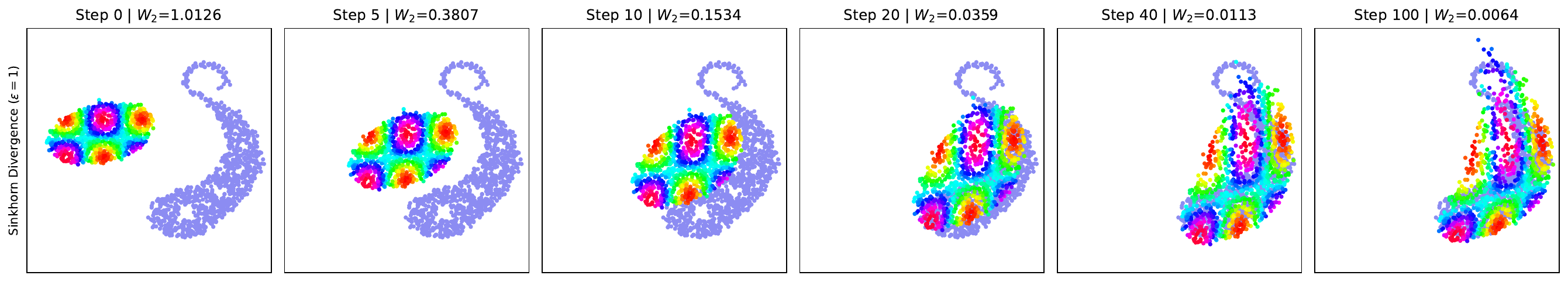} \\
\includegraphics[width=1\linewidth]{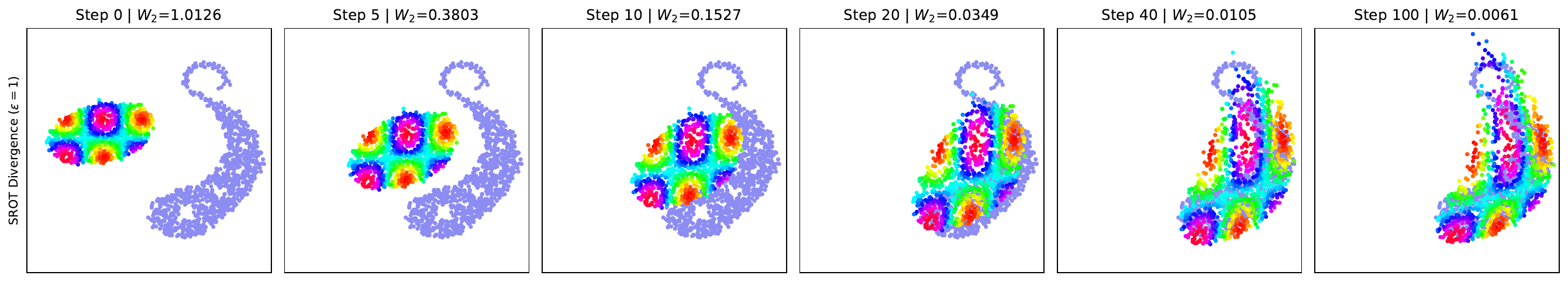} 
        
    \end{tabular}
}
    \caption{\footnotesize{Gradient flows of Sinkhorn divergence and SR divergence with Wasserstein distance as netural evaluation metric.}}
    \label{fig:gd_appendix}
    \vspace{-0.5em}
\end{figure}

\textbf{Gradient flow.} We report the result of the gradient flow from Sinkhorn divergence and SROT divergence with  $\varepsilon =1$ in Figure~\ref{fig:gd_appendix}. Overall, we observe that SROT is slightly better than EOT.

\begin{figure}[!t]
\centering
\includegraphics[width=0.7\linewidth]{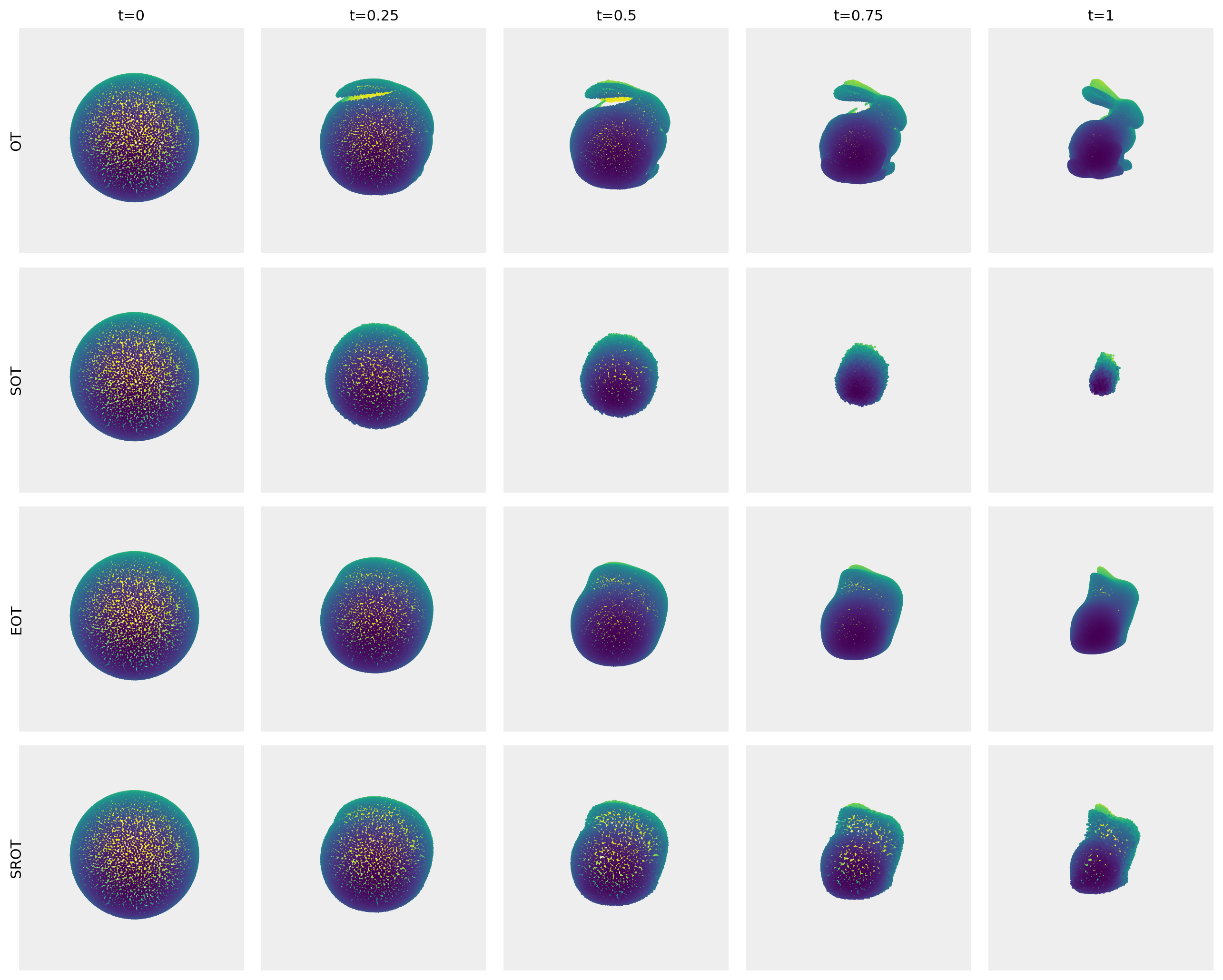}\\[2pt]
{\small (a) Sphere to Stanford bunny}\\[8pt]
\includegraphics[width=0.7\linewidth]{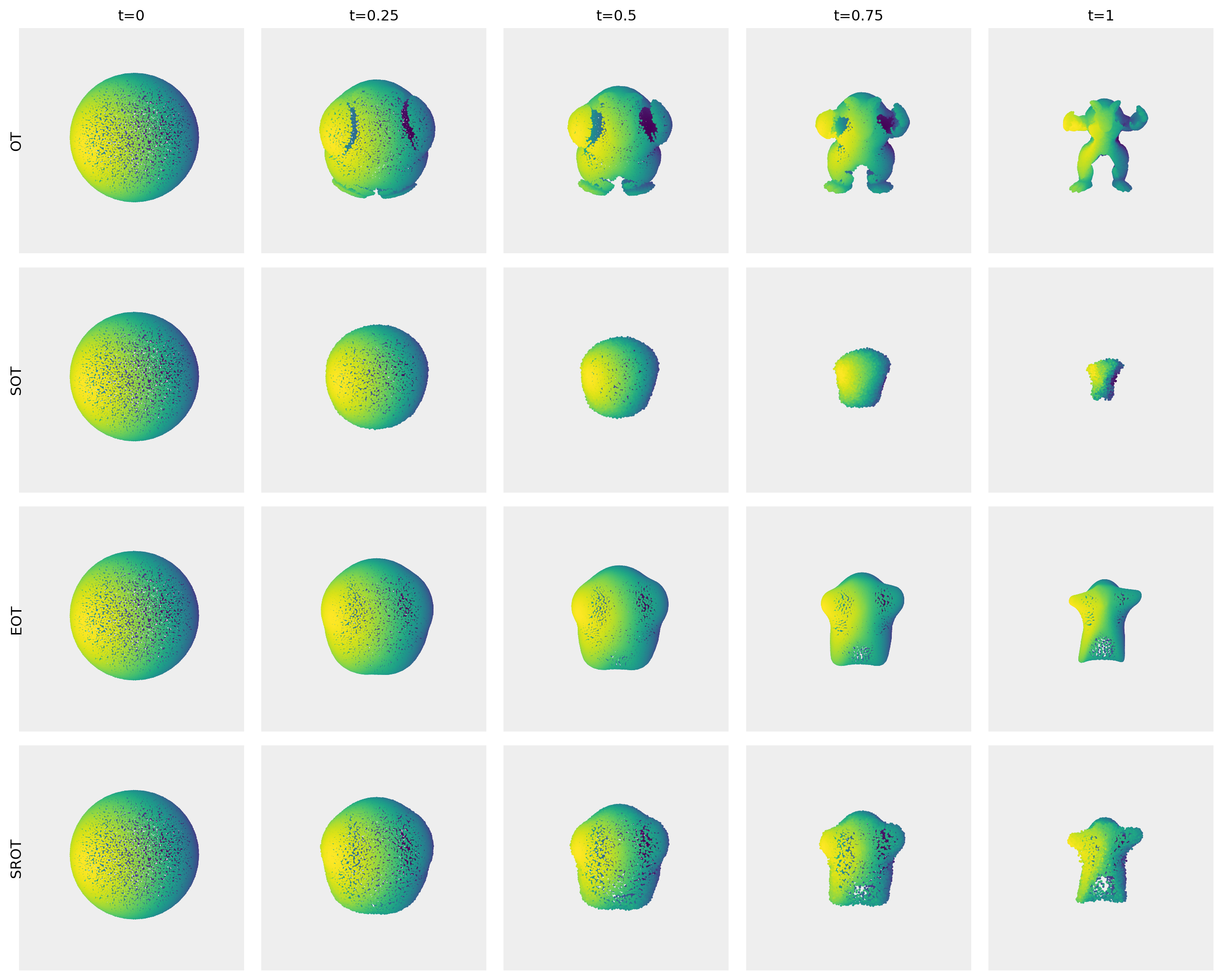}\\[2pt]
{\small (b) Sphere to Armadillo}
\caption{Displacement interpolation between two point clouds of $10{,}000$ points
under the OT, SOT, EOT, and SROT ($\gamma = 0$) plans ($\varepsilon = 10^{-2}$,
$L = 100$). Each source point moves to its barycentric image under the plan.}
\label{fig:mesh_interp}
\end{figure}

{\textbf{Transport between 3D Point Clouds.}
We also explore the problem of transporting point clouds in $\mathbb{R}^3$. The source is
$10{,}000$ points uniformly sampled from the unit sphere, and the target is $10{,}000$
points sampled from either the Armadillo or the Stanford bunny using the Poisson disk
sampling algorithm. There meshes are centered and scaled  to the unit ball,
with the same squared Euclidean cost normalized by its maximum value. We compare the
performance of SOT, EOT, and SROT with $\gamma = 10^{-8}$ and $\gamma = 0$, taking
$L = 100$ directions, $\varepsilon \in \{1, 10^{-1}, 10^{-2}\}$, running at most
$T = 5000$ Sinkhorn iterations, with a tolerance of $10^{-9}$ on the marginal
constraints, and reporting $\mathbb{L}_1$ distance, relative transportation cost gap to
the exact plan and wall-clock time. These metrics are computed over the average of the two meshes
(Table~\ref{tab:mesh_sparse}). Figure~\ref{fig:mesh_interp} shows the displacement
interpolation from the sphere to each mesh, where every source point travels along a
straight line path to its barycentric image in the transportation plan. SROT lowers
the cost gap of EOT for all values of $\varepsilon$, and SROT with $\gamma = 0$ is
the fastest algorithm, because SOT supports only one percent of the $n \times m$
cells.  In the interpolation, OT deforms the sphere into the mesh,
the entropic plans reach a smoothed version of it, SROT keeping more of the
structure than EOT, and the SOT plan alone collapses the cloud.
}
\begin{table}[!t]
\centering
\caption{$\mathbb{L}_1$ error, relative transportation cost gap and wall-clock time
versus exact OT (mean over the two meshes) for transporting a sphere to the
Armadillo and to the Stanford bunny ($n=m=10{,}000$ points, $L=100$, $T=5000$).}
\label{tab:mesh_sparse}
\scalebox{0.8}{
\begin{tabular}{l ccc ccc ccc}
\toprule
 & \multicolumn{3}{c}{$\epsilon = 1$} & \multicolumn{3}{c}{$\epsilon = 0.1$} & \multicolumn{3}{c}{$\epsilon = 0.01$} \\
\cmidrule(lr){2-4} \cmidrule(lr){5-7} \cmidrule(lr){8-10}
Method & $\mathbb{L}_1$ & gap & time (s) & $\mathbb{L}_1$ & gap & time (s) & $\mathbb{L}_1$ & gap & time (s) \\
\midrule
SOT & \multicolumn{9}{c}{$\mathbb{L}_1 = 1.999$, \; gap $= +236\%$, \; time $= 0.26$ s \;(independent of $\epsilon$)} \\
\midrule
EOT & 2.000 & $+328\%$ & 1.91 & 1.999 & $+129\%$ & 1.87 & 1.991 & $+13.8\%$ & 5.10 \\
SROT ($\gamma=10^{-8}$) & 1.999 & $\mathbf{+205\%}$ & 2.45 & 1.998 & $\mathbf{+80\%}$ & 2.44 & 1.991 & $+12.7\%$ & 6.88 \\
SROT ($\gamma=0$) & 1.999 & $\mathbf{+205\%}$ & \textbf{0.42} & 1.998 & $\mathbf{+80\%}$ & \textbf{0.44} & 1.991 & $\mathbf{+12.5\%}$ & \textbf{0.71} \\
\bottomrule
\end{tabular}}
\end{table}

\clearpage

\bibliography{example_paper}
\bibliographystyle{abbrv}

\end{document}